\documentclass{article}
\PassOptionsToPackage{numbers,sort&compress}{natbib}
\usepackage[preprint]{neurips_2025}
\makeatletter
\renewcommand{\@notice}{\enlargethispage{2\baselineskip}}
\makeatother
\usepackage[utf8]{inputenc}
\usepackage[T1]{fontenc}
\usepackage{amsmath,amssymb}
\usepackage{graphicx}
\usepackage{placeins}
\usepackage{booktabs,array}
\usepackage{microtype}
\usepackage{xcolor}
\usepackage{tikz}
\usetikzlibrary{arrows.meta,positioning}
\usepackage{url}
\usepackage{hyperref}
\hypersetup{hidelinks,pdftitle={Torch-DPhys: A Torch-Native, MOOSE-Compatible Framework for Differentiable Physics Simulation and Hybrid Machine Learning Integration},pdfauthor={Akzhol Almukhametov and Yang Liu}}

\title{Torch-DPhys: A Torch-Native, MOOSE-Compatible Framework for Differentiable Physics Simulation and Hybrid Machine Learning Integration}
\author{%
\begin{tabular}[b]{@{}c@{\hspace{4em}}c@{}}
\textbf{Akzhol Almukhametov} & \textbf{Yang Liu}\\[0.4em]
\normalfont Department of Nuclear Engineering & \normalfont Department of Nuclear Engineering\\
\normalfont Texas A\&M University & \normalfont Texas A\&M University\\[0.4em]
\normalfont\texttt{akzhol\_almukhametov@tamu.edu} & \normalfont\texttt{y-liu@tamu.edu}
\end{tabular}}

\begin{document}
\maketitle
\raggedbottom

\begin{abstract}
Gradient-based parameter inference, optimal control, and solver-in-the-loop hybrid physics--machine learning (ML) require derivatives of simulation objectives with respect to model parameters, boundary conditions, or embedded neural network weights. While existing nuclear engineering codes provide mature multiphysics forward solvers, coupling them to modern automatic differentiation (AD) and machine learning ecosystems typically requires external cross-language wrappers or code-intrusive adjoint implementations. We present Torch-DPhys, a PyTorch/LibTorch-native framework for differentiable physics simulation and hybrid physics--ML integration that implements selected modeling concepts and discretizations from the Multiphysics Object-Oriented Simulation Environment (MOOSE) Thermal Hydraulics Module (THM), the System Analysis Module (SAM), and finite-element continuum mechanics. Torch-DPhys expresses finite-element and finite-volume residuals as native tensor operations, supporting execution on central processing units (CPUs) and graphics processing units (GPUs), kernel compilation via \texttt{torch.compile}, and sparse direct linear solvers through SuperLU and NVIDIA's cuDSS library for direct sparse solvers. Simulation sensitivities are computed via two complementary routes: reverse-mode differentiation unrolled through executed solver iterations, and discrete residual adjoints evaluated at converged implicit states. Forward numerical verification against analytical solutions, MOOSE, and SAM-Lite demonstrates close discrete agreement across plane-strain elasticity, one-dimensional single- and two-loop thermal-hydraulic transients, and two-dimensional axisymmetric molten-salt fast reactor (MSFR) flow and heat transport, where Torch-DPhys matches SAM-Lite within a maximum normalized field discrepancy of $1.13\times 10^{-5}$ over 200 adaptive time steps. Transient trajectory gradients are verified against tangent-linear recurrences and central finite differences. We demonstrate the framework on two gradient-driven applications: recovering pump head and heating power in a thermal-hydraulic loop within $0.06\%$ relative error from synthetic MOOSE observations, and training a recurrent neural controller via differentiable predictive control (DPC) by backpropagating closed-loop trajectory objectives through a three-loop plant solver. Torch-DPhys connects MOOSE- and SAM-compatible physical modeling with PyTorch/LibTorch's native differentiation ecosystem, enabling gradient-based inverse modeling, optimal control, and solver-in-the-loop hybrid physics--machine learning for nuclear engineering systems.
\end{abstract}

\section{Introduction}

Numerical simulation in nuclear science and engineering is traditionally formulated as a forward problem: given a system geometry, material properties, boundary conditions, and empirical closure relations, a solver integrates the governing conservation equations to predict thermal-hydraulic or thermomechanical states. In many engineering workflows, however, forward prediction is embedded within an outer optimization or inference loop. Modelers frequently need to infer uncertain boundary or constitutive parameters from sparse sensor measurements, optimize component geometries or operating schedules, or synthesize feedback controllers that satisfy safety constraints during transients. Addressing these tasks via partial differential equation (PDE)-constrained optimization requires sensitivities that relate trajectory-level objectives to the underlying physical parameters or control actions. Because finite-difference approximations and derivative-free search methods scale poorly as the parameter dimension grows, access to accurate simulation gradients via adjoint methods or automatic differentiation (AD) is essential.

The demand for solver gradients has grown as data assimilation and hybrid physics--machine learning (ML) have become central to nuclear system modeling and digital twin development~\cite{liu2022samml,abulawi2025bayesian,lim2026facility,11683640}. Early hybrid workflows coupled ML with mechanistic solvers offline: neural networks were trained on pre-generated high-fidelity simulation databases to predict turbulent eddy viscosity or closure terms and were subsequently linked into system codes for forward deployment~\cite{liu2022samml,liu2024msrsamml}. Similarly, control-oriented studies have often trained control policies by backpropagating through learned surrogate models, such as neural ordinary differential equations (ODEs), rather than the physical plant solver itself~\cite{lim2026facility,11683640}. While computationally convenient, decoupling training from the numerical solver introduces distribution shift and surrogate approximation error; a closure or controller that performs well offline can degrade in accuracy or stability when deployed in closed loop with the discretized governing equations. Propagating gradients directly through the time-stepping physics solver—known as solver-in-the-loop training~\cite{farrell2013,fan2024}—allows neural closures and control policies to be optimized against multi-step trajectory objectives while accounting for the numerical discretization and physical coupling of the solver.

Within the nuclear engineering community, the Multiphysics Object-Oriented Simulation Environment (MOOSE)~\cite{harbour2025} and its domain applications, including the Thermal Hydraulics Module (THM)~\cite{hansel2024} and the System Analysis Module (SAM)~\cite{hu2021sam,hu2025sam,hu2026sammsr}, provide widely adopted infrastructure for reactor thermal-hydraulics and multiphysics analysis. MOOSE employs forward-mode AD to assemble local elemental and nodal residual Jacobians~\cite{lindsay2021} and supports forward evaluation of trained neural networks via C++ LibTorch bindings~\cite{german2023}. However, because these production codes are architected in C++ around data structures distributed using the Message Passing Interface (MPI) and nonlinear solvers from the Portable, Extensible Toolkit for Scientific Computation (PETSc), they do not natively expose end-to-end reverse-mode differentiation of transient trajectory objectives directly within Python-based ML training loops.

In the broader scientific computing literature, two primary strategies have emerged to bridge numerical PDE solvers and differentiable programming. The first strategy wraps an established compiled solver with an external adjoint interface. For example, Diff-NekRS~\cite{jung2026diffnekrs} retains NekRS for forward spectral-element fluid simulation, implements a manual discrete adjoint for the supported time-stepping scheme, and couples the resulting sensitivities to LibTorch for multi-step neural closure training. The second strategy implements the numerical discretizations natively within tensor-computing frameworks such as JAX, PyTorch, or TensorFlow. Representative examples include JAX-FEM~\cite{xue2023jaxfem} for differentiable finite-element mechanics and PhiFlow~\cite{holl2024phiflow} for differentiable fluid simulation. Native tensor implementation places the physical residual evaluations, linear solvers, and neural network layers within a single execution and differentiation environment~\cite{paszke2019}, avoiding cross-language serialization while supporting both unrolled computational-graph differentiation and implicit adjoint formulations.

In this work, we develop Torch-DPhys, a PyTorch-native framework for differentiable physics simulation and hybrid physics--ML integration tailored to nuclear thermal-hydraulic and continuum mechanics modeling. Rather than replacing full production codes, Torch-DPhys implements selected physical components and finite-element/finite-volume discretizations derived from MOOSE THM~\cite{hansel2024} and SAM~\cite{hu2021sam,hu2025sam} alongside two-dimensional finite-element elasticity. Component topologies and mesh operators are lowered into structured tensor representations that execute on CPUs or NVIDIA GPUs, with support for kernel compilation via \texttt{torch.compile} and sparse direct linear algebra via SuperLU~\cite{demmel1999superlu,li1999superluguide} and NVIDIA cuDSS~\cite{nvidia2026cudss}. To accommodate different solver structures and memory budgets, Torch-DPhys provides two routes to trajectory gradients: \emph{(i)} unrolled reverse-mode AD through executed nonlinear solver iterations, and \emph{(ii)} discrete residual adjoints evaluated on the converged implicit equations at each time step.

The main contributions of this work are as follows:
\begin{itemize}
    \item \textbf{A PyTorch-native multiphysics modeling architecture.} We formulate object-oriented physical models—spanning plane-strain linear elasticity, one-dimensional (1D) compressible channel and loop thermal hydraulics with conjugate heat structures and junctions, and two-dimensional (2D) axisymmetric stabilized finite-element flow and heat transport—as native PyTorch tensor operations that run in the same environment as ML optimization workflows.
    \item \textbf{Dual differentiation routes with transient gradient verification.} We implement both unrolled Newton-iteration differentiation (with step-wise checkpoint recomputation) and converged-residual discrete adjoints (including variable-step second-order backward differentiation (BDF2) history terms), and verify the resulting sensitivities against independent tangent-linear recurrences and central finite differences.
    \item \textbf{Cross-code numerical verification against MOOSE and SAM-Lite.} We verify forward simulations against archived MOOSE results across static elasticity and three 1D thermal-hydraulic loop configurations, and against SAM-Lite on a 200-step adaptive transient of an axisymmetric molten-salt fast reactor (MSFR), achieving a maximum normalized field discrepancy of $1.13\times 10^{-5}$ while characterizing time-step stabilization sensitivities and mass-flux conservation properties.
    \item \textbf{Evaluation of sparse compiled CPU and GPU execution.} Using the 200-step MSFR transient workload, we benchmark eager and \texttt{torch.compile}-accelerated execution across CPU threads and GPU hardware (with graphs in NVIDIA's CUDA parallel computing platform and cuDSS sparse factorizations), analyzing component-level costs across residual evaluation, Jacobian assembly, sparse factorization, and adjoint dependency discovery.
    \item \textbf{Demonstrations of inverse parameter estimation and solver-in-the-loop control.} We apply Torch-DPhys to recover pump head and heating power in a thermal-hydraulic loop from sparse transient MOOSE observations (within $0.06\%$ relative error), and to train a gated recurrent unit (GRU) control policy via differentiable predictive control (DPC) through a three-loop thermal-hydraulic plant model.
\end{itemize}

The remainder of this paper is organized as follows. Section~\ref{sec:framework} describes the Torch-DPhys software architecture and the two routes for computing simulation gradients. Section~\ref{sec:models} summarizes the governing equations and discretizations for the structural mechanics, 1D channel thermal-hydraulics, and 2D axisymmetric MSFR models. Section~\ref{sec:verification} presents forward verification against MOOSE and SAM-Lite, transient sensitivity verification, and time-step refinement diagnostics. Section~\ref{sec:performance} analyzes CPU and GPU execution performance on the MSFR workload. Section~\ref{sec:applications} demonstrates PDE-constrained parameter identification, solver-in-the-loop DPC policy training, and extensions to uncertainty quantification. Section~\ref{sec:conclusion} discusses current limitations and concludes the paper.

\section{Framework Design and Differentiation}
\label{sec:framework}

\subsection{Physical Models as Torch Computations}

Torch-DPhys organizes materials, boundary conditions, and domain-specific physical components as Python objects while evaluating their local constitutive relations and global residual assembly through PyTorch tensor operations. Across the supported physics modules—two-dimensional finite-element mechanics, one-dimensional system thermal hydraulics, the reduced three-loop control plant, and two-dimensional axisymmetric finite-element flow and heat transport—this design separates high-level model specification from vectorized numerical execution.

For the structural mechanics module, mesh generation and finite-element basis preprocessing are performed offline using the MFEM modular finite-element library~\cite{mfem2021}. MFEM reads or generates the computational mesh, constructs first-order triangular finite-element spaces, and exports nodal coordinates, element connectivity, physical basis-function gradients, quadrature weights, and boundary node indices. Torch-DPhys loads these static arrays at initialization and executes stiffness operator evaluations and conjugate-gradient solves natively in PyTorch without runtime calls to MFEM.

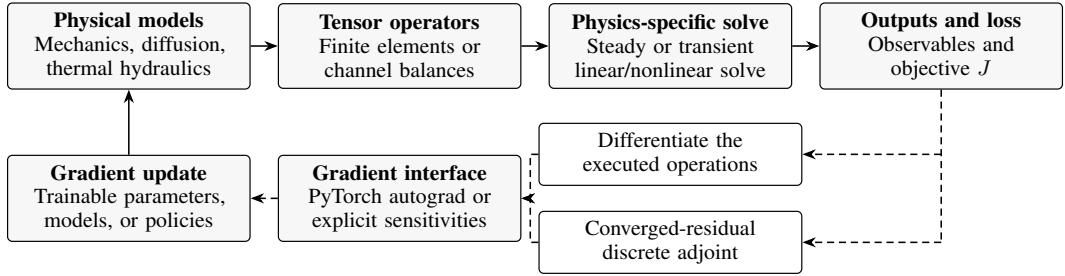
\begin{figure}[ht!]
    \centering
    \resizebox{\linewidth}{!}{%
\begin{tikzpicture}[
        font=\small,
        block/.style={draw, rounded corners=2pt, line width=0.5pt,
            fill=black!3, align=center, text width=3.3cm,
            minimum height=1.25cm, inner sep=4pt},
        route/.style={block, fill=white, text width=3.6cm,
            minimum height=0.9cm},
        forward/.style={-{Stealth[length=2mm]}, line width=0.65pt},
        backward/.style={forward, densely dashed}
    ]
        % Forward model: no assumption that lowering is mandatory.
        \node[block] (model) at (0,0)
            {\textbf{Physical models}\\Mechanics, diffusion,\\thermal hydraulics};
        \node[block] (operators) at (4,0)
            {\textbf{Tensor operators}\\Finite elements or\\channel balances};
        \node[block] (solve) at (8,0)
            {\textbf{Physics-specific solve}\\Steady or transient\\linear/nonlinear solve};
        \node[block] (loss) at (12,0)
            {\textbf{Outputs and loss}\\Observables and\\objective $J$};
        \draw[forward] (model) -- (operators);
        \draw[forward] (operators) -- (solve);
        \draw[forward] (solve) -- (loss);

        % Alternative derivative implementations, exposed through one interface.
        \node[route] (graph) at (8,-1.6)
            {Differentiate the\\executed operations};
        \node[route] (adjoint) at (8,-2.9)
            {Converged-residual\\discrete adjoint};
        \node[block] (autograd) at (4,-2.25)
            {\textbf{Gradient interface}\\PyTorch autograd or\\explicit sensitivities};
        \node[block] (update) at (0,-2.25)
            {\textbf{Gradient update}\\Trainable parameters,\\models, or policies};
        \draw[densely dashed, line width=0.65pt]
            (loss.south) -- (12,-2.9);
        \draw[backward] (12,-1.6) -- (graph.east);
        \draw[backward] (12,-2.9) -- (adjoint.east);
        \draw[backward] (graph.west) -- (5.93,-1.6) |- (autograd.east);
        \draw[backward] (adjoint.west) -- (5.93,-2.9) |- (autograd.east);
        \draw[backward] (autograd) -- (update);
        \draw[forward] (update.north) -- (model.south);
    \end{tikzpicture}}
    \caption{Torch-DPhys workflow. Solid arrows show forward evaluation and parameter updates. Dashed arrows show the two differentiation routes: unrolled differentiation through executed solver operations and discrete adjoints of the converged residual equations.}
    \label{fig:architecture}
\end{figure}

For the one-dimensional thermal-hydraulic models, a lowering stage translates the object-oriented component network into contiguous tensor index tables representing channel connectivity, junction topology, heat-structure coupling, and physical parameters. This lowering step reformulates the component equations for batched tensor evaluation without altering the underlying physical conservation laws. Residual evaluations, automatic differentiation, and linear algebra execute on either CPU or NVIDIA CUDA devices, and computationally intensive residual and Jacobian kernels can be compiled using \texttt{torch.compile}. Figure~\ref{fig:architecture} illustrates the overall execution and differentiation workflow.

\subsection{Two Routes to Simulation Gradients}
\label{sec:differentiation_routes}

Torch-DPhys computes gradients of a scalar objective $J$ with respect to model parameters $\boldsymbol\theta$ via two distinct mathematical routes, which can be exposed either through PyTorch's native autograd engine or through an explicit sensitivity interface (as in the MSFR module):

\begin{enumerate}
    \item \textbf{Unrolled differentiation of executed operations.} This route applies reverse-mode AD directly to the sequence of numerical operations executed during the forward solve, such as a fixed number of conjugate-gradient steps or damped Newton iterations. Because the computational graph traces the actual solver updates, the resulting derivative represents the exact sensitivity of the finite-iteration algorithm, including dependence on the initial guess, for the executed sequence of solver updates. However, retaining intermediate solver states across long transients increases memory consumption unless step-wise checkpointing and recomputation are used, and gradients from partially converged iterations need not match the sensitivity of the exact implicit root.

    \item \textbf{Discrete adjoint of the converged residual equations.} Rather than differentiating through the nonlinear solver path, the residual-adjoint route applies the implicit function theorem to the converged discrete equations $\mathbf R_n(\mathbf y_n, \mathbf y_{n-1}, \boldsymbol\theta) = \mathbf 0$ at each time step $n$. Assuming a differentiable residual and a nonsingular state Jacobian $\partial\mathbf R_n/\partial\mathbf y_n$, the discrete adjoint $\boldsymbol\lambda_n$ satisfies the linear system
    \begin{equation}
     \left(\frac{\partial\mathbf R_n}{\partial\mathbf y_n}\right)^T
     \boldsymbol\lambda_n=\overline{\mathbf y}_n,
     \qquad
     \Delta\nabla_{\boldsymbol\theta}J=
     -\left(\frac{\partial\mathbf R_n}{\partial\boldsymbol\theta}\right)^T
     \boldsymbol\lambda_n,
     \label{eq:adjoint}
    \end{equation}
    where $\overline{\mathbf y}_n$ combines the direct objective derivative $\partial J/\partial\mathbf y_n$ with history contributions propagated backward from step $n+1$ (and $n+2$ for two-step schemes). Explicit dependence of $J$ on $\boldsymbol\theta$ is added separately. The residual adjoint requires solving only one transposed linear system per accepted time step and avoids storing intermediate Newton iterates, though its accuracy depends on how tightly both the forward residual and adjoint linear system are converged.
\end{enumerate}

Section~\ref{sec:parameter_identification} compares both routes directly on a transient thermal-hydraulic inverse problem, while Section~\ref{sec:hybrid} uses adjoint-based trajectory sensitivities to train a closed-loop neural control policy. Detailed mathematical formulations of both differentiation targets are provided in Appendix~\ref{app:differentiation}.

\section{Representative Physical Models}
\label{sec:models}

To demonstrate the framework across distinct spatial discretizations and physics regimes, we implement three representative model families in PyTorch: two-dimensional finite-element linear elasticity, one-dimensional finite-volume system thermal hydraulics, and two-dimensional axisymmetric stabilized finite-element flow and heat transport.

\subsection{Structural Mechanics}

The structural mechanics module solves two-dimensional isotropic, small-strain linear elasticity under plane-strain conditions:
\begin{equation}
 -\nabla\cdot\boldsymbol\sigma=\mathbf b,\qquad
 \boldsymbol\epsilon=\tfrac12(\nabla\mathbf u+\nabla\mathbf u^T),\qquad
 \boldsymbol\sigma=\lambda\,\mathrm{tr}(\boldsymbol\epsilon)\mathbf I+2\mu\boldsymbol\epsilon,
 \label{eq:elasticity}
\end{equation}
where $\mathbf u$ is the displacement vector, $\mathbf b$ is the body force per unit volume, $\boldsymbol\epsilon$ is the infinitesimal strain tensor, $\boldsymbol\sigma$ is the Cauchy stress tensor, and $\lambda$ and $\mu$ are the Lam\'e parameters. Using precomputed MFEM basis gradients on linear triangular elements, tensor operators gather nodal displacements, evaluate element strains and stresses at quadrature points, and scatter-add internal nodal forces. This matrix-free operator evaluation computes the action of the global stiffness matrix directly within a preconditioned conjugate-gradient solver while symmetric boundary projections enforce Dirichlet constraints.

\subsection{THM-Derived Channel Thermal Hydraulics}

The one-dimensional thermal-hydraulic module implements single-phase compressible flow and conjugate heat transfer following the formulation of the MOOSE Thermal Hydraulics Module~\cite{hansel2024}. Within a flow channel of cross-sectional area $A$, conservation of mass, axial momentum, and total energy is governed by
\begin{equation}
 \mathbf U=\begin{bmatrix}\rho A\\\rho uA\\\rho EA\end{bmatrix},\qquad
 \mathbf F=\begin{bmatrix}\rho uA\\(\rho u^2+p)A\\u(\rho E+p)A\end{bmatrix},\qquad
 \frac{d\mathbf U_i}{dt}=-\frac{\widehat{\mathbf F}_{i+1/2}-\widehat{\mathbf F}_{i-1/2}}{\Delta x}+\mathbf S_i,
 \label{eq:flow}
\end{equation}
where $\rho$, $u$, and $p$ denote fluid density, axial velocity, and static pressure, $E = e + u^2/2$ is the specific total energy, and $\mathbf S_i$ collects body-force, wall-friction, and wall heat-transfer source terms. Thermodynamic properties are closed via fluid equations of state. Inter-cell numerical fluxes $\widehat{\mathbf F}_{i\pm 1/2}$ are evaluated using the Harten--Lax--van Leer--contact (HLLC) approximate Riemann solver~\cite{toro1994} with linear interface reconstruction.

Flow channels are coupled into system networks through volume junctions, parallel-channel junctions, and pump components. Conjugate heat structures solve radial or transverse solid heat conduction and exchange thermal power with adjacent fluid cells via the convective flux relation $q' = h P_h (T_s - T_f)$, where $h$ is the convective heat-transfer coefficient, $P_h$ is the heated perimeter, and $T_s$ and $T_f$ are the solid surface and bulk fluid temperatures. Assembling the fluid, junction, solid conduction, and controller states into a global state vector $\mathbf y$ yields the semi-discrete system $\dot{\mathbf y} = \mathbf f(\mathbf y, \boldsymbol\theta, t)$, which is integrated in time using the implicit backward Euler scheme:
\begin{equation}
 \mathbf R_n=\frac{\mathbf y_n-\mathbf y_{n-1}}{\Delta t_n}
 -\mathbf f(\mathbf y_n,\boldsymbol\theta,t_n)=\mathbf0.
 \label{eq:residual}
\end{equation}
Equation~\eqref{eq:residual} defines both the forward Newton residual and the implicit state constraint differentiated in Eq.~\eqref{eq:adjoint}.

\subsection{SAM-Derived Axisymmetric Flow and Heat Transport}
\label{sec:msfr_model}

To evaluate Torch-DPhys on multidimensional finite-element fluid and thermal transport, we implement the two-dimensional axisymmetric porous-medium incompressible Navier--Stokes and energy equations from SAM~\cite{hu2021sam,hu2026sammsr}. We apply this model to the Molten Salt Fast Reactor (MSFR) configuration studied within the EURATOM EVOL project~\cite{rouch2014msfr,brovchenko2019msfrbenchmark}, following the axisymmetric SAM setup introduced by Liu et al.~\cite{liu2024msrsamml}. In the present verification study, we model an open-cavity core with unit porosity, zero porous resistance, and a standard Prandtl mixing-length algebraic turbulence closure during a heated startup transient with a fixed inlet velocity profile.

\begin{figure}[htbp]
 \centering
 \reflectbox{\includegraphics[angle=90,width=0.42\linewidth]{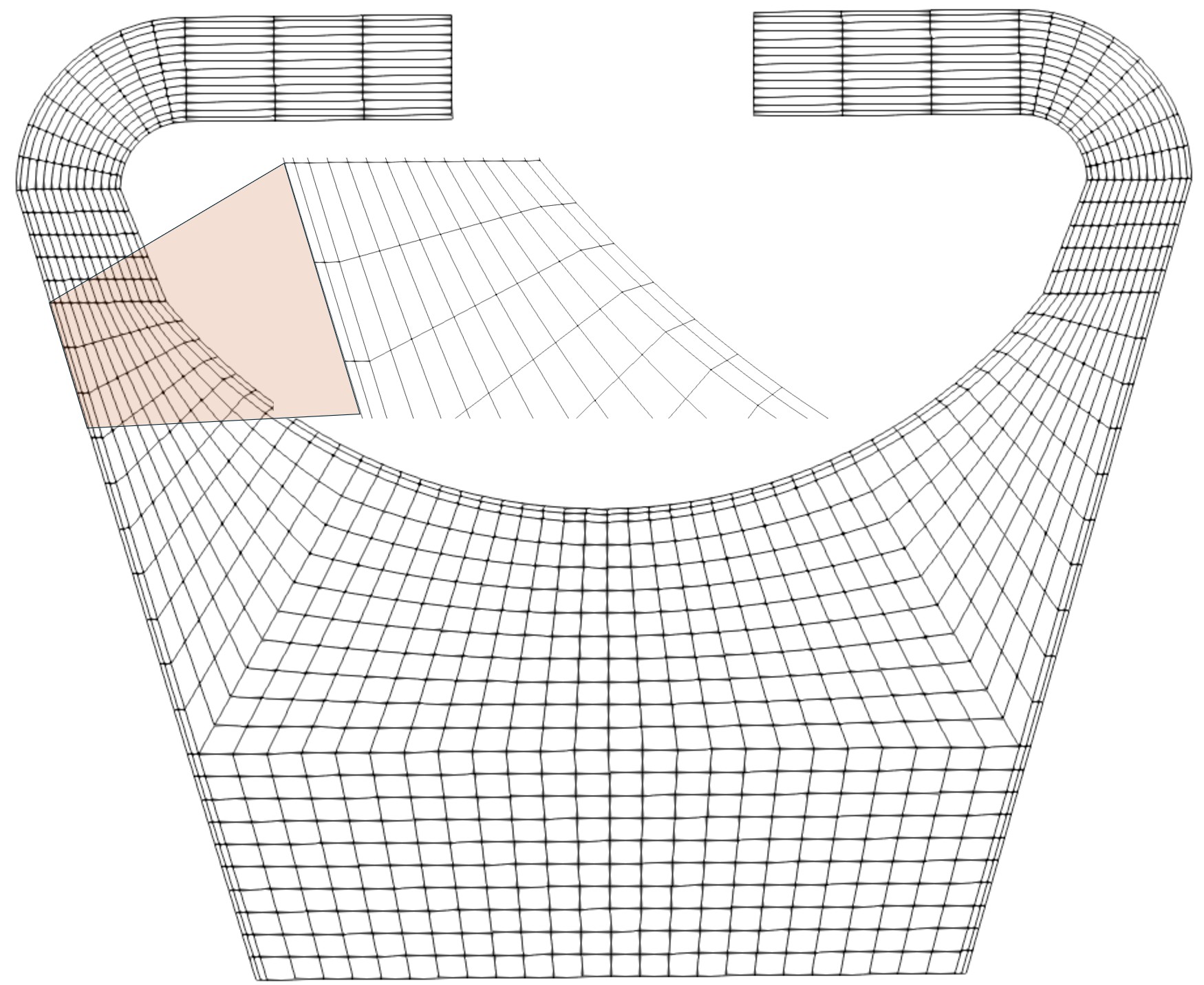}}
 \caption{Coarse mesh of the axisymmetric MSFR computational domain, with enlarged detail. The axial direction is vertical and the symmetry axis is on the left.}
 \label{fig:msfr_mesh}
\end{figure}

Figure~\ref{fig:msfr_mesh} shows the axisymmetric computational domain and coarse mesh.

The axisymmetric domain is discretized into 1,296 curved nine-node isoparametric quadrilateral elements (`QUAD9`), with bilinear (`Q1`) solution spaces defined on the 1,399 corner vertices for axial velocity $u_z$, radial velocity $u_r$, pressure $p$, and temperature $T$ (5,596 total degrees of freedom). Time integration uses a backward-Euler startup step followed by variable-step second-order backward differentiation (BDF2). Table~\ref{tab:msfr_model_settings} summarizes the thermophysical properties, mixing-length profile, and boundary conditions.

\begin{table}[htbp]
 \centering
 \small
 \caption{Physical settings of the MSFR initialization benchmark. Coordinates $x$ and $r$ denote the axial and radial directions.}
 \label{tab:msfr_model_settings}
 \begin{tabular}{@{}>{\raggedright\arraybackslash}p{0.28\linewidth}>{\raggedright\arraybackslash}p{0.67\linewidth}@{}}
 \toprule
 Quantity & Configuration \\
 \midrule
 Fluid properties & Constant $\rho=4306.71$~kg/m$^3$, $c_p=1593.94$~J/(kg\,K), and $\mu=0.0253665$~Pa\,s. Molecular conductivity varies linearly from 0.9952 at 800~K to 1.0288~W/(m\,K) at 1200~K. \\
 Mixing length & 0.07~m for $r<1.25$~m, decreasing linearly to 0.018~m at $r=1.75$~m, and 0.018~m beyond. Turbulent Prandtl number $Pr_t=0.9$. \\
 Inlet & Recorded normalized Nek5000 velocity profile, multiplied by $-1.25$~m/s and the coarse-boundary flux correction 1.02705297222. Radial velocity is zero and temperature is 900~K. \\
 Outlet, walls and axis & Outlet pressure 161~kPa. Walls are no-slip and adiabatic. The symmetry axis has zero radial velocity and zero normal gradients of axial velocity and temperature. \\
 Initial fields & Both velocity components $10^{-5}$~m/s, pressure 110~kPa and temperature 900~K, subject to the prescribed boundary values. \\
 Body force and heating & Gravity is $+9.81$~m/s$^2$ in the mesh axial direction. The fixed volumetric heat source is given by Eq.~\eqref{eq:msfr_heat_source}. \\
 \bottomrule
 \end{tabular}
\end{table}

The volumetric fission heat source $q'''(x,r)$ is sampled at the finite-element nodes and interpolated across quadrature points using the clipped cosine profile
\begin{equation}
 q'''(x,r)=q_0+q_1
 \left[\cos\!\left(\frac{\pi\sqrt{x^2+r^2}}{2a}\right)\right]_+
 \left[\cos\!\left(\frac{\pi|x|}{2b}\right)\right]_+,
 \label{eq:msfr_heat_source}
\end{equation}
where $[z]_+=\max(0,z)$, $q_0=3.0\times10^6$~W/m$^3$, $q_1=6.19\times10^8$~W/m$^3$, $a=1.13319021$~m, and $b=1.35099968$~m.

Because equal-order bilinear interpolation is used for velocity and pressure in a convection-dominated regime, the weak form incorporates Pressure-Stabilizing/Petrov--Galerkin (PSPG) stabilization in the continuity equation and Streamline-Upwind/Petrov--Galerkin (SUPG) stabilization in the momentum and energy equations~\cite{hu2021sam}. For an element $e$ with minimum characteristic length $h_e$ and quadrature-averaged velocity $\overline{\mathbf u}_e$, the stabilization parameters are defined as
\begin{align}
 A_e &= \frac{4}{\Delta t^2}+\frac{4\|\overline{\mathbf u}_e\|^2}{h_e^2},
 & \tau_c &= A_e^{-1/2}, \label{eq:msfr_tau_base}\\
 \tau_m &= \left[A_e+\left(\frac{4\nu_{\mathrm{eff}}}{h_e^2}\right)^2\right]^{-1/2},
 & \tau_e &= \left[A_e+\left(\frac{4\alpha}{h_e^2}\right)^2\right]^{-1/2},
 \label{eq:msfr_tau_transport}
\end{align}
where $\nu_{\mathrm{eff}}=(\mu+\mu_t)/\rho$ and $\alpha=k/(\rho c_p)$ are evaluated at quadrature points from the eddy viscosity $\mu_t$ and molecular conductivity $k$, while the diffusive energy flux uses the element-averaged total conductivity $k + c_p \mu_t / Pr_t$. Because $\Delta t$ enters $A_e$ directly in Eq.~\eqref{eq:msfr_tau_base}, changing the time step modifies the spatial stabilization weights $\tau_c$, $\tau_m$, and $\tau_e$ even as the flow approaches steady state; we analyze the consequences of this coupling in Section~\ref{sec:msfr_stage1_benchmark}.

\section{Numerical Verification}
\label{sec:verification}

We verify the forward discretizations and sensitivity calculations in Torch-DPhys against analytical solutions, archived MOOSE simulations, and SAM-Lite reference runs, followed by tangent-linear and finite-difference gradient checks. Note that agreement between Torch-DPhys and a reference code under matched meshes and time steps verifies the equivalence of the discrete implementations, whereas spatial and temporal continuum convergence must be assessed separately (Section~\ref{sec:msfr_stage1_benchmark}).

\subsection{Structural Mechanics Verification}

We first verify the plane-strain elasticity module on two static benchmark problems using linear triangular elements with Young's modulus $E=1000$ and Poisson's ratio $\nu=0.3$ (in consistent dimensionless units): an 81-node unit-square patch test with prescribed affine boundary displacements, and an 85-node $4\times 1$ cantilever beam clamped at its left boundary and subjected to a uniform downward body force $\mathbf b = (0, -1)$.

\begin{figure}[ht!]
 \centering
 \includegraphics[width=\linewidth]{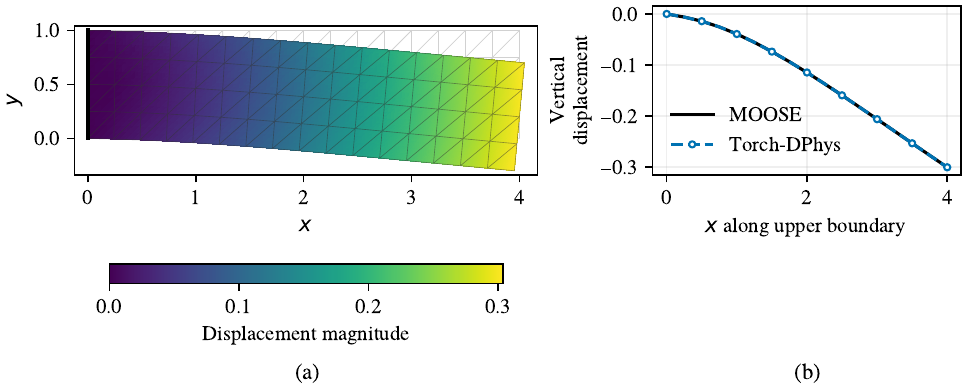}
 \caption{Plane-strain cantilever verification. (a) Deformed mesh colored by displacement magnitude (unmagnified). (b) Upper-boundary vertical displacement profile compared against archived MOOSE results.}
 \label{fig:mechanics}
\end{figure}

On the patch test, Torch-DPhys recovers the exact analytical linear displacement field and matches the archived MOOSE solution with a maximum componentwise relative $L_2$ discrepancy of $1.6\times 10^{-14}$. On the cantilever beam (Figure~\ref{fig:mechanics}), the deformed mesh and upper-boundary vertical displacement profile match MOOSE with a maximum relative $L_2$ discrepancy of $2.4\times 10^{-13}$.

\FloatBarrier
\subsection{Thermal Hydraulics: Heated Inlet-Driven Circuit}
\label{sec:upper03}

\begin{figure}[htbp]
 \centering
 \includegraphics[width=\linewidth]{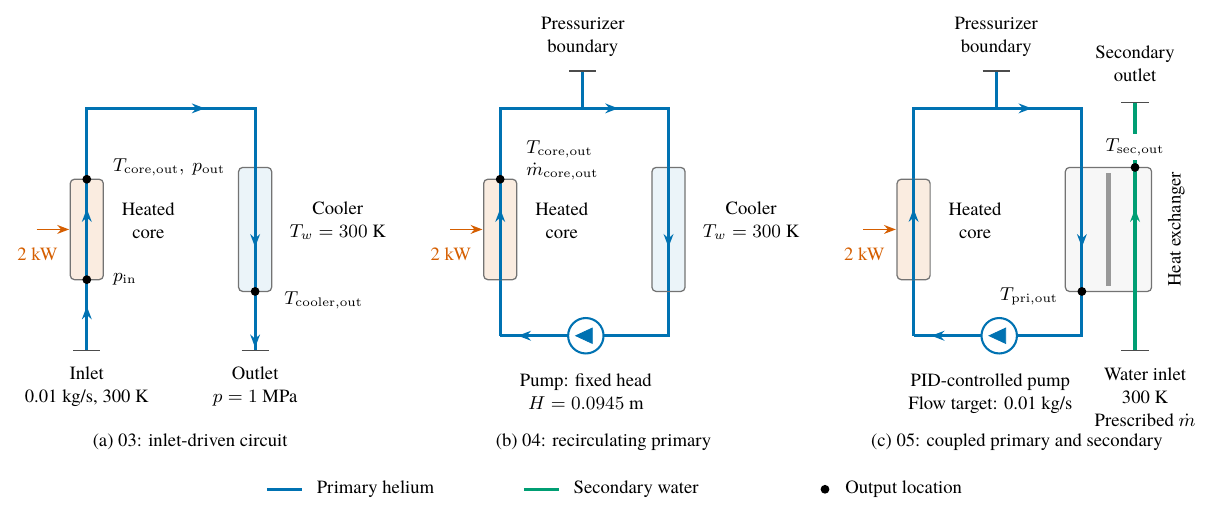}
 \caption{Schematics of the one-dimensional thermal-hydraulic verification cases: (a) Case 03 open helium chain, (b) Case 04 closed loop with constant pump head, and (c) Case 05 coupled primary--secondary system with PID control. Arrows indicate nominal flow direction and dots mark output sensor locations.}
 \label{fig:thm_cases}
\end{figure}

Figure~\ref{fig:thm_cases} illustrates the three MOOSE THM test configurations used to verify the one-dimensional thermal-hydraulic module: (a) an open inlet-driven helium heating and cooling chain (\texttt{03\_upper\_loop.i}), (b) a closed recirculating primary loop driven by a constant-head pump (\texttt{04\_loop.i} variant), and (c) a coupled primary--secondary heat-exchanger system with proportional--integral--derivative (PID) pump control (\texttt{05\_secondary\_side.i}).

\FloatBarrier

Case \texttt{03\_upper\_loop.i} (Figure~\ref{fig:thm_cases}a) tests conjugate fluid--solid heat transfer and parallel-channel junction coupling (\texttt{JunctionParallelChannels1Phase}) across seven connected helium flow channels with gravity disabled. Helium enters at $0.01$~kg/s and $300$~K against a $1$~MPa outlet pressure, passes through a cylindrical core structure heated at $2000$~W, and exits through a downstream cooler with walls maintained at $300$~K. Starting from an initial velocity of $10^{-4}$~m/s, Torch-DPhys advances the transient using backward Euler and a dense Newton solver with backtracking line search across the nine archived MOOSE time steps up to $500$~s (as well as an eight-step reduced heated-section variant up to $250$~s), converging on every interval without time-step cutbacks.

\begin{figure}[htbp]
 \centering
 \includegraphics[width=\linewidth]{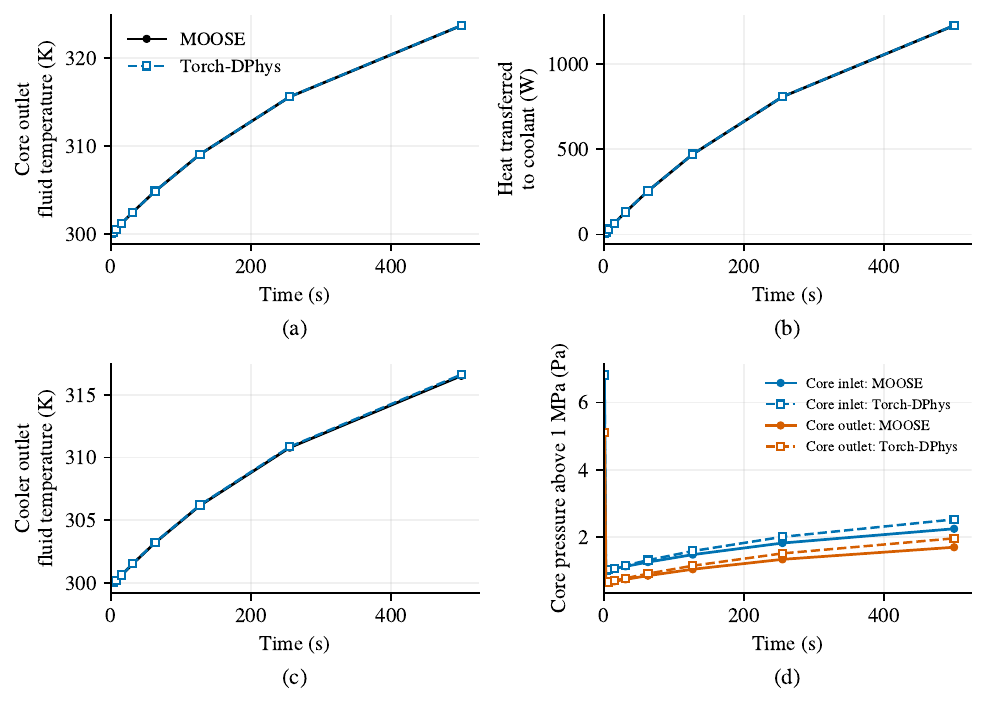}
 \caption{Comparison between Torch-DPhys and MOOSE for the \texttt{03\_upper\_loop.i} configuration: (a) core outlet fluid temperature, (b) heat transfer rate from the solid core to the coolant, (c) cooler outlet temperature, and (d) core inlet and outlet pressures relative to $1$~MPa.}
 \label{fig:upper03}
\end{figure}

\begin{figure}[htbp]
 \centering
 \includegraphics[width=\linewidth]{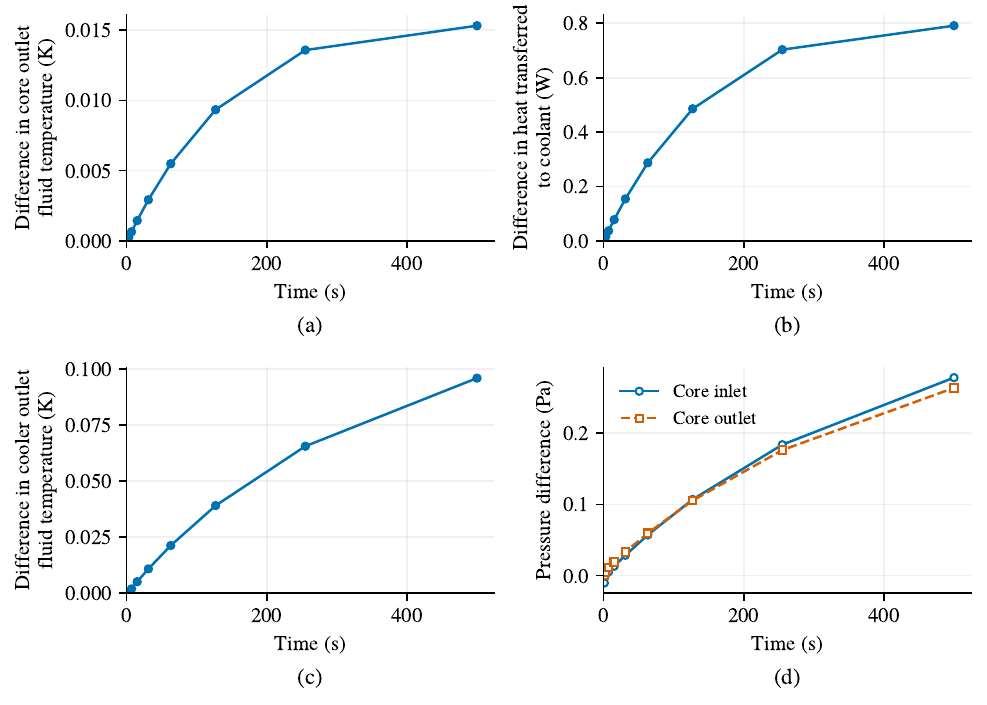}
 \caption{Signed differences (Torch-DPhys minus MOOSE) corresponding to the four observables in Figure~\ref{fig:upper03}.}
 \label{fig:upper03diff}
\end{figure}

Figures~\ref{fig:upper03} and~\ref{fig:upper03diff} compare the Torch-DPhys and MOOSE trajectories and plot their signed differences across all nine positive time steps ($t \in [1, 500]$~s). Across the full circuit, the maximum absolute discrepancies between Torch-DPhys and MOOSE are $0.0153$~K in core outlet temperature, $0.0958$~K in cooler outlet temperature, $0.791$~W in solid-to-coolant heat transfer rate, and $0.278$~Pa and $0.264$~Pa in core inlet and outlet pressures (relative to the $1$~MPa baseline). On the reduced eight-step heated-section problem, the corresponding maximum differences are $0.0135$~K, $0.699$~W, $0.180$~Pa, and $0.173$~Pa.

\subsection{Thermal Hydraulics: Primary and Coupled Two-Loop Systems}

\begin{figure}[ht!]
 \centering
 \includegraphics[width=\linewidth]{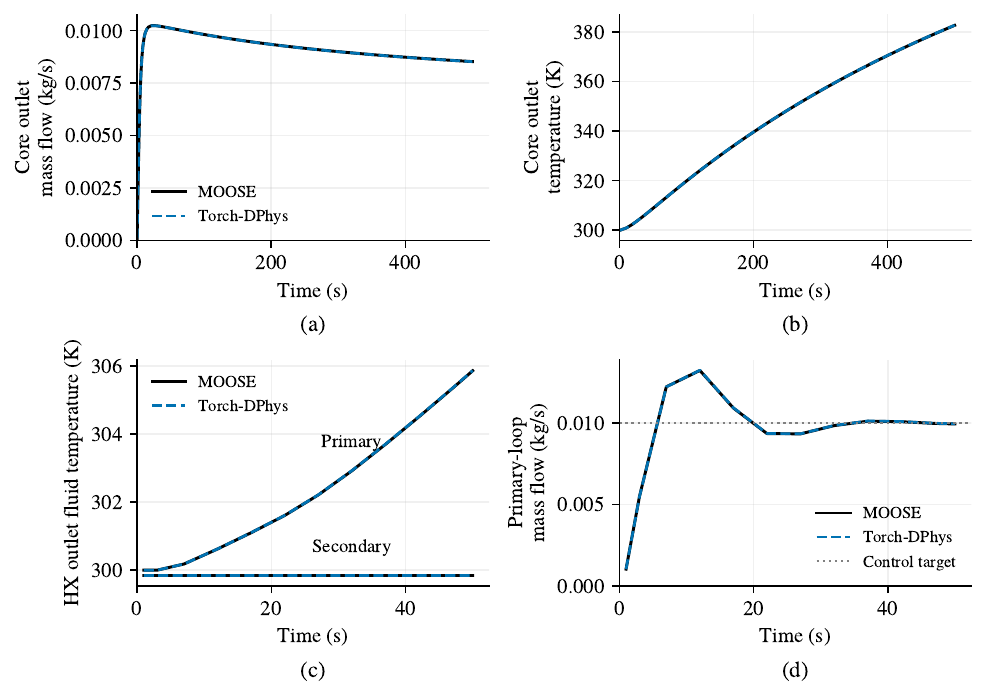}
 \caption{Comparisons between Torch-DPhys and MOOSE for the closed primary loop (\texttt{04\_loop.i} with constant pump head): (a) core outlet mass flow and (b) core outlet temperature; and for the coupled two-loop system (\texttt{05\_secondary\_side.i}): (c) primary and secondary heat-exchanger outlet temperatures and (d) PID-controlled primary-loop mass flow rate.}
 \label{fig:forward04}
\end{figure}

Figure~\ref{fig:forward04} compares Torch-DPhys against MOOSE on the closed primary loop (\texttt{04\_loop.i} variant, panels a--b) and the PID-controlled coupled primary--secondary loop (\texttt{05\_secondary\_side.i}, panels c--d).

In the \texttt{04\_loop.i} configuration (Figure~\ref{fig:thm_cases}b), a constant-head pump circulates fluid through a heated core and pipe network connected by volume junctions, with gravity and feedback control disabled. Both codes step through the same 122 prescribed non-uniform time steps over $500$~s. Despite minor discretization differences—Torch-DPhys uses unlimited monotonic upstream-centered scheme for conservation laws (MUSCL) interface reconstruction whereas the MOOSE reference uses a minmod limiter, along with minor differences in radial solid meshing and junction momentum initialization—the trajectories agree closely across the transient, with maximum absolute differences of $2.24\times 10^{-6}$~kg/s in core outlet mass flow rate (Figure~\ref{fig:forward04}a) and $0.041$~K in core outlet fluid temperature (Figure~\ref{fig:forward04}b).

In the \texttt{05\_secondary\_side.i} configuration (Figure~\ref{fig:thm_cases}c), the primary loop transfers heat through a counterflow heat exchanger (HX) to a once-through secondary channel while a PID controller regulates the primary pump head to maintain a target primary mass flow rate of $0.01$~kg/s. Over the 12 prescribed time steps up to $50$~s, the maximum differences between Torch-DPhys and MOOSE at the primary and secondary HX outlets are $0.00687$~K and $1.41\times 10^{-5}$~K, respectively (Figure~\ref{fig:forward04}c), and remain below $0.017$~K across all four monitored temperature locations. The primary mass flow rate (Figure~\ref{fig:forward04}d) captures the initial overshoot and settling toward the $0.01$~kg/s setpoint with a maximum discrepancy of $1.76\times 10^{-5}$~kg/s ($0.133\%$ of the peak flow rate, reflecting the comparison between a channel-cell flow estimate in Torch-DPhys and a junction-flux postprocessor in MOOSE), while the core pressure drop agrees within $0.00144$~Pa.

\subsection{MSFR Forward Verification Against SAM-Lite}
\label{sec:sam_residual}

We verify the two-dimensional axisymmetric MSFR implementation against SAM-Lite at both the residual evaluation level and across full transient trajectories. At eight archived reference states, the assembled Torch-DPhys residual vector matches SAM-Lite bit-for-bit at every degree of freedom except at four inlet boundary nodes in four states, where floating-point multiplication order in the inlet velocity expression introduces a one-unit-in-the-last-place (1~ULP) difference. Component-level finite-difference checks of the automatic state Jacobian, BDF2 history derivatives, and parameter Jacobians are documented in Appendix~\ref{app:sam_diagnostics}.

For transient forward verification under the baseline input deck tolerances ($10^{-6}$ relative and absolute scaled residual norms), Torch-DPhys forms the exact residual Jacobian via AD and solves each Newton step using a direct lower--upper (LU) factorization with cubic backtracking, whereas SAM-Lite uses preconditioned Jacobian-free Newton--Krylov (PJFNK). Starting from the same initial state, both solvers independently select identical 200-step adaptive time grids reaching $t = 43.7148$~s without step rejections, typically requiring 1--3 nonlinear iterations per step.

\begin{figure}[htbp]
 \centering
 \includegraphics[width=\linewidth]{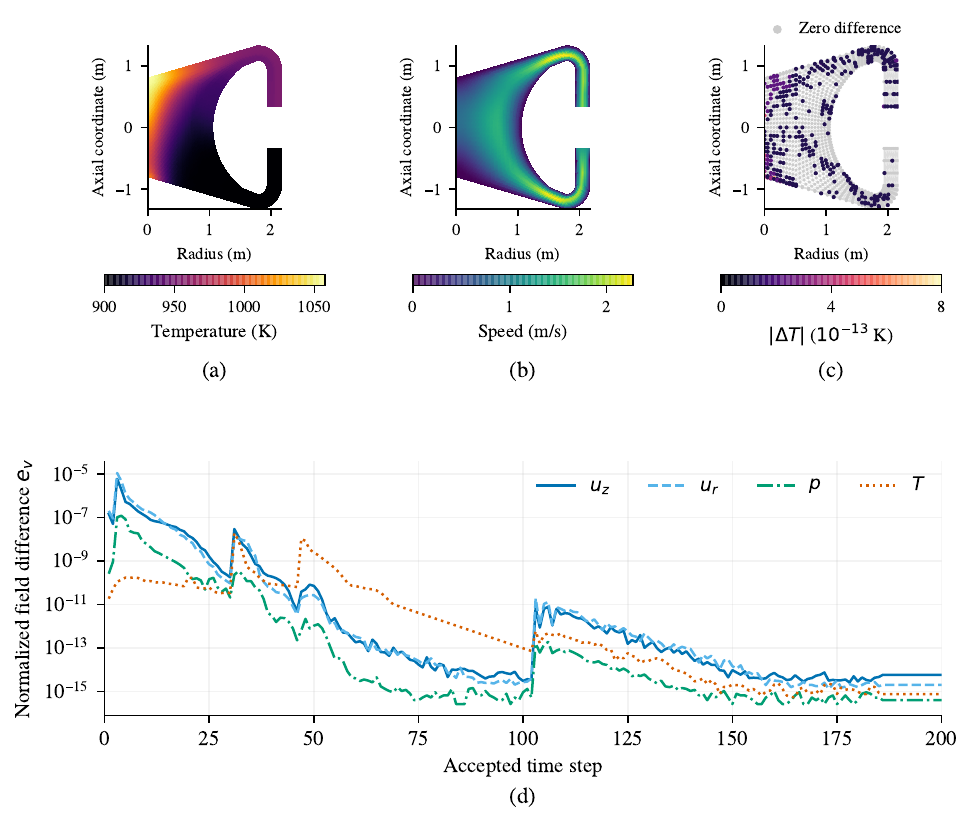}
 \caption{MSFR forward verification over the 200-step adaptive transient. (a,b) Reference SAM-Lite temperature and velocity magnitude fields at $t=43.7$~s. (c) Nodal absolute temperature difference $|T^{\mathrm{Torch}} - T^{\mathrm{SAM}}|$ at $t=43.7$~s in units of $10^{-13}$~K (gray indicates exact bitwise zero). (d) Evolution of the normalized field discrepancies $e_v(k)$ defined in Eq.~\eqref{eq:msfr_field_error} over all 200 steps.}
 \label{fig:msfr_fields}
\end{figure}

Figure~\ref{fig:msfr_fields} and Table~\ref{tab:msfr_absolute_fields} summarize the field-level agreement over the 200-step transient. To compare all four solution fields ($u_z, u_r, p, T$) on a common dimensionless scale in Figure~\ref{fig:msfr_fields}(d), we define the step-wise normalized $L_\infty$ discrepancy for field $v$ at step $k$ as
\begin{equation}
 e_v(k)=\frac{\max_i|u^{\mathrm{Torch}}_{v,i}(k)-u^{\mathrm{SAM}}_{v,i}(k)|}
 {\max_i|u^{\mathrm{SAM}}_{v,i}(k)|},
 \label{eq:msfr_field_error}
\end{equation}
where $i \in \{1,\ldots,1399\}$ indexes the mesh vertices. Across all 200 steps, the maximum normalized discrepancy is $1.13\times 10^{-5}$ (corresponding to maximum physical differences of $1.94\times 10^{-5}$~K in temperature at step 31 and $2.85\times 10^{-2}$~Pa in pressure at step 4), and drops below $7.1\times 10^{-11}$ from step 60 onward. At the final state ($t = 43.7$~s), the nodal temperature fields agree to within $7.96\times 10^{-13}$~K, with 1,016 of 1,399 vertices matching bit-for-bit. Across the eager and compiled CPU and GPU backends evaluated in Section~\ref{sec:msfr_200}, eager and compiled CPU states are bitwise identical at every step, and GPU trajectories differ from CPU by at most $1.6\times 10^{-14}$ under Eq.~\eqref{eq:msfr_field_error}.

\begin{table}[htb]
 \centering
 \caption{Maximum absolute nodal discrepancies between Torch-DPhys and SAM-Lite across all 1,399 vertices and steps 1--200 of the MSFR initialization transient.}
 \label{tab:msfr_absolute_fields}
 \begin{tabular}{lcrr}
 \toprule
 Field & Unit & Maximum absolute difference & Step\\
 \midrule
 Temperature $T$ & K & $1.94\times10^{-5}$ & 31\\
 Axial velocity $u_z$ & m/s & $1.02\times10^{-5}$ & 3\\
 Radial velocity $u_r$ & m/s & $2.28\times10^{-5}$ & 3\\
 Pressure $p$ & Pa & $2.85\times10^{-2}$ & 4\\
 \bottomrule
 \end{tabular}
\end{table}

\subsection{MSFR Transient-Gradient Verification}
\label{sec:msfr_sensitivities}

For a sequence of accepted implicit states satisfying the two-step BDF2 residual $\mathbf R_k(\mathbf y_k,\mathbf y_{k-1},\mathbf y_{k-2},\boldsymbol\theta)=\mathbf0$ on a fixed time-step schedule, let $\mathbf A_k=\partial\mathbf R_k/\partial\mathbf y_k$. Given a scalar trajectory objective $L$ and fixed initial conditions, the discrete residual adjoint backward recurrence is
\begin{equation}
 \begin{aligned}
 \mathbf A_k^T\boldsymbol\lambda_k
 &=\nabla_{\mathbf y_k}L
 -\left(\frac{\partial\mathbf R_{k+1}}{\partial\mathbf y_k}\right)^T\boldsymbol\lambda_{k+1}
 -\left(\frac{\partial\mathbf R_{k+2}}{\partial\mathbf y_k}\right)^T\boldsymbol\lambda_{k+2},\\
 \mathbf g_{\mathrm{traj}}
 &=-\sum_{k=1}^{N_{\mathrm{acc}}}\left(\frac{\partial\mathbf R_k}{\partial\boldsymbol\theta}\right)^T\boldsymbol\lambda_k,
 \qquad
 \nabla_{\boldsymbol\theta}L=\partial_{\boldsymbol\theta}L+\mathbf g_{\mathrm{traj}},
 \end{aligned}
 \label{eq:msfr_bdf2_adjoint}
\end{equation}
where boundary terms with indices $k+1 > N_{\mathrm{acc}}$ or $k+2 > N_{\mathrm{acc}}$ vanish, and backward-Euler steps omit the second-history term. When a trajectory contains rejected time-step attempts, the BDF2 history metadata (\texttt{dt\_old}) from the retry sequence are held fixed while the adjoint recurrence steps backward across the accepted states.

To verify that every state variable and time step contributes accurately to $\mathbf g_{\mathrm{traj}}$ without masking errors behind sparse sensor locations, we define a dense verification functional over all $N_{\mathrm{acc}}$ accepted states:
\begin{equation}
 L_{\mathrm{ver}}=\sum_{k=1}^{N_{\mathrm{acc}}}\langle\mathbf W_k,\mathbf U_k\rangle_F
 =\sum_{k=1}^{N_{\mathrm{acc}}}\sum_{a=1}^{4}\sum_{i=1}^{1399}
 W_{k,a,i}U_{k,a,i},
 \label{eq:msfr_verification_objective}
\end{equation}
where $\mathbf U_k\in\mathbb R^{4\times1399}$ stacks the nodal velocity, pressure, and temperature fields at step $k$, and $\mathbf W_k$ is a fixed array of seeded standard-normal weights.

\begin{figure}[htbp]
 \centering
 \includegraphics[width=0.85\linewidth]{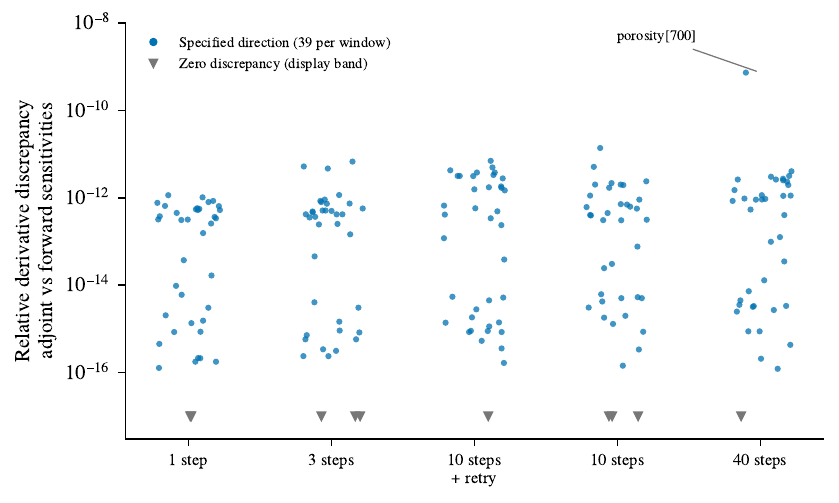}
 \caption{Relative discrepancy between the MSFR discrete residual adjoint and forward tangent-linear sensitivities across 39 parameter directions on trajectory windows of 1 to 40 accepted steps. Triangles in the lower band indicate exact bitwise zero discrepancy; ``+ retry'' denotes the 10-step window containing a rejected step attempt.}
 \label{fig:msfr_adjoint}
\end{figure}

We verify the discrete adjoint~\eqref{eq:msfr_bdf2_adjoint} against an independently implemented forward tangent-linear sensitivity recurrence (Eq.~\eqref{eq:msfr_forward_sensitivity} in Appendix~\ref{app:msfr_adjoint}) and central finite differences. Figure~\ref{fig:msfr_adjoint} shows the relative directional discrepancy $|a-b|/\max(|a|,|b|)$ between the adjoint and forward-sensitivity directional derivatives across 39 parameter directions on five trajectory windows ranging from 1 to 40 accepted steps (including a 10-step window that replays a recorded step-9 rejection and retry). Across all 39 directions and five windows, the maximum relative discrepancy is $7.37\times 10^{-10}$.

Extending the gradient verification to the full 200-step transient on the compiled CPU trajectory, the discrete adjoint agrees with the tangent-linear recurrence to within $4.2\times 10^{-13}$ across four tested directions (inlet temperature, inlet velocity scale, volumetric heat-source amplitude, and a random direction spanning all 34 registered parameter blocks). Across the four execution backends (eager and compiled CPU and GPU), the 200-step adjoint gradients agree within a maximum blockwise relative infinity-norm difference of $2.7\times 10^{-12}$. Furthermore, central finite-difference checks performed on a tightly converged 200-step trajectory (relative tolerance $10^{-12}$) match the adjoint derivatives to within $3.7\times 10^{-8}$ at a relative perturbation size of $10^{-4}$ (see Appendix~\ref{app:msfr_adjoint} for conditioning and tolerance-polishing diagnostics).

\subsection{Time-Step Sensitivity and Conservation Diagnostics}
\label{sec:msfr_stage1_benchmark}

Although Torch-DPhys and SAM-Lite agree to within $1.13\times 10^{-5}$ on the 200-step adaptive schedule, uniform time-step refinement of the production MSFR formulation reveals a notable sensitivity. Refining a constant integration step from $\Delta t = 0.25$~s to $0.03125$~s over $t \in [2.71, 43.7]$~s shifts the endpoint mean outlet temperature by $+0.46$~K, maximum core temperature by $+0.8$~K, and mean inlet pressure by $+2.2$~kPa ($5.8\%$ of the total inlet--outlet pressure drop), while the global mass-flux imbalance between the integrated inlet and outlet velocity profiles increases from $1.65\%$ to $1.90\%$ (Table~\ref{tab:msfr_timestep_sensitivity} in Appendix~\ref{app:msfr_timestep}).

This behavior stems from the explicit appearance of $\Delta t$ in the SUPG/PSPG stabilization coefficient $A_e = 4/\Delta t^2 + 4\|\overline{\mathbf u}_e\|^2/h_e^2$ in Eqs.~\eqref{eq:msfr_tau_base}--\eqref{eq:msfr_tau_transport}: reducing $\Delta t$ decreases $\tau_c$, $\tau_m$, and $\tau_e$, which alters the stabilized spatial weak form even at steady state. When $\Delta t$ inside the stabilization coefficients is held fixed at $0.25$~s while the BDF2 time-integration step is refined from $0.25$~s to $0.03125$~s, the solutions converge monotonically toward a common endpoint state with an observed temporal order of $1.90\text{--}2.15$ for temperature. This diagnostic confirms that the time-step shift and coarse-mesh mass-flux imbalance are inherent properties of the stabilized finite-element formulation shared by SAM-Lite and Torch-DPhys rather than discrepancies in the Torch implementation.

\section{Execution and Computational Performance}
\label{sec:performance}

We use the 200-step axisymmetric MSFR transient and its 200-step discrete adjoint to evaluate the computational performance of eager and compiled CPU and GPU execution in Torch-DPhys alongside serial and MPI-parallelized SAM-Lite.

\subsection{Sparse Linear Algebra, Compilation, and GPU Execution}
\label{sec:msfr_cost}
\label{sec:msfr_gpu}

At each Newton iteration and adjoint step, the 5,596-degree-of-freedom MSFR problem requires solving a sparse linear system whose Jacobian contains 182,809 nonzeros ($0.58\%$ fill). To handle the wide dynamic range of the unscaled matrix entries (condition number $\sim 5.5\times 10^{16}$), Torch-DPhys applies row equilibration followed by sparse direct LU factorization and one step of iterative refinement (Appendix~\ref{app:msfr_linear_algebra}). On the CPU, sparse factorizations use SciPy's SuperLU backend~\cite{demmel1999superlu,li1999superluguide} with column approximate minimum degree (COLAMD) ordering; on the GPU, factorizations use NVIDIA cuDSS~\cite{nvidia2026cudss} invoked through a custom C++ PyTorch operator. Topology-dependent sparsity maps are precomputed once and reused across all steps.

In compiled execution, \texttt{torch.compile} generates fused kernels for the finite-element residual, vector--Jacobian products (VJPs), and element-Jacobian assembly, while Python retains outer Newton and time-stepping control. On the GPU, compiled kernels are captured within CUDA graphs to reduce host launch overhead, and all state vectors, sparse matrix values, and adjoint history records remain resident in device memory throughout the transient. Moreover, an auxiliary-variable reuse pass evaluates intermediate constitutive and stabilization quantities once per accepted adjoint step rather than repeating them across dependency checks and Jacobian preparation, preserving bitwise-identical gradients while reducing adjoint overhead.

\subsection{200-Step CPU/GPU Performance Benchmark}
\label{sec:msfr_200}

Table~\ref{tab:msfr_benchmark_setup} summarizes the hardware and software environment for the 200-step MSFR performance benchmark. Across all Torch-DPhys configurations (eager and compiled, CPU and GPU), the adaptive solver executes the exact same 200 time steps, 266 Newton iterations, 467 residual evaluations, 280 Jacobian assemblies, and 266 sparse LU factorizations.

\begin{table}[!htbp]
 \centering
 \caption{Hardware and software configuration for the 200-step MSFR performance benchmark.}
 \label{tab:msfr_benchmark_setup}
 \small
 \renewcommand{\arraystretch}{1.15}
 \begin{tabular}{@{}p{0.22\linewidth}p{0.73\linewidth}@{}}
 \toprule
 Item & Setting\\
 \midrule
 Workload & 200 adaptive forward steps to $t=43.7148$~s at baseline tolerances (266 Newton iterations), followed by a 200-step discrete residual adjoint.\\
 Hardware and precision & Intel Core Ultra 7 265K CPU, NVIDIA RTX 2000 Ada GPU (16~GB), \texttt{float64} arithmetic under Windows Subsystem for Linux 2 (WSL2).\\
 Software stack & Python 3.12.3, PyTorch 2.13.0+cu129, Triton 3.7.1, CUDA 12.9, cuDSS 0.8.0.10, NumPy 2.1.2, SciPy 1.16.1, Intel oneAPI Math Kernel Library (MKL) 2024.2.\\
 Thread and rank settings & Torch-DPhys: 4, 8, or 16 Torch/OpenMP/MKL threads (with \texttt{MKL\_CBWR=AUTO,STRICT} and 1 OpenBLAS thread). SAM-Lite: 1 rank with 1 or 4 MOOSE threads, or 4, 8, and 16 MPI ranks with 1 thread per rank.\\
 Measurement protocol & Medians and min--max ranges over three warm repetitions for Torch-DPhys; single runs for cold compilation and SAM-Lite. Forward timings measure \texttt{Transient.execute()} (Torch-DPhys) and cumulative \texttt{PicardSolve} (SAM-Lite).\\
 \bottomrule
 \end{tabular}
\end{table}

\begin{figure}[htbp]
 \centering
 \includegraphics[width=\linewidth,height=0.82\textheight,keepaspectratio]{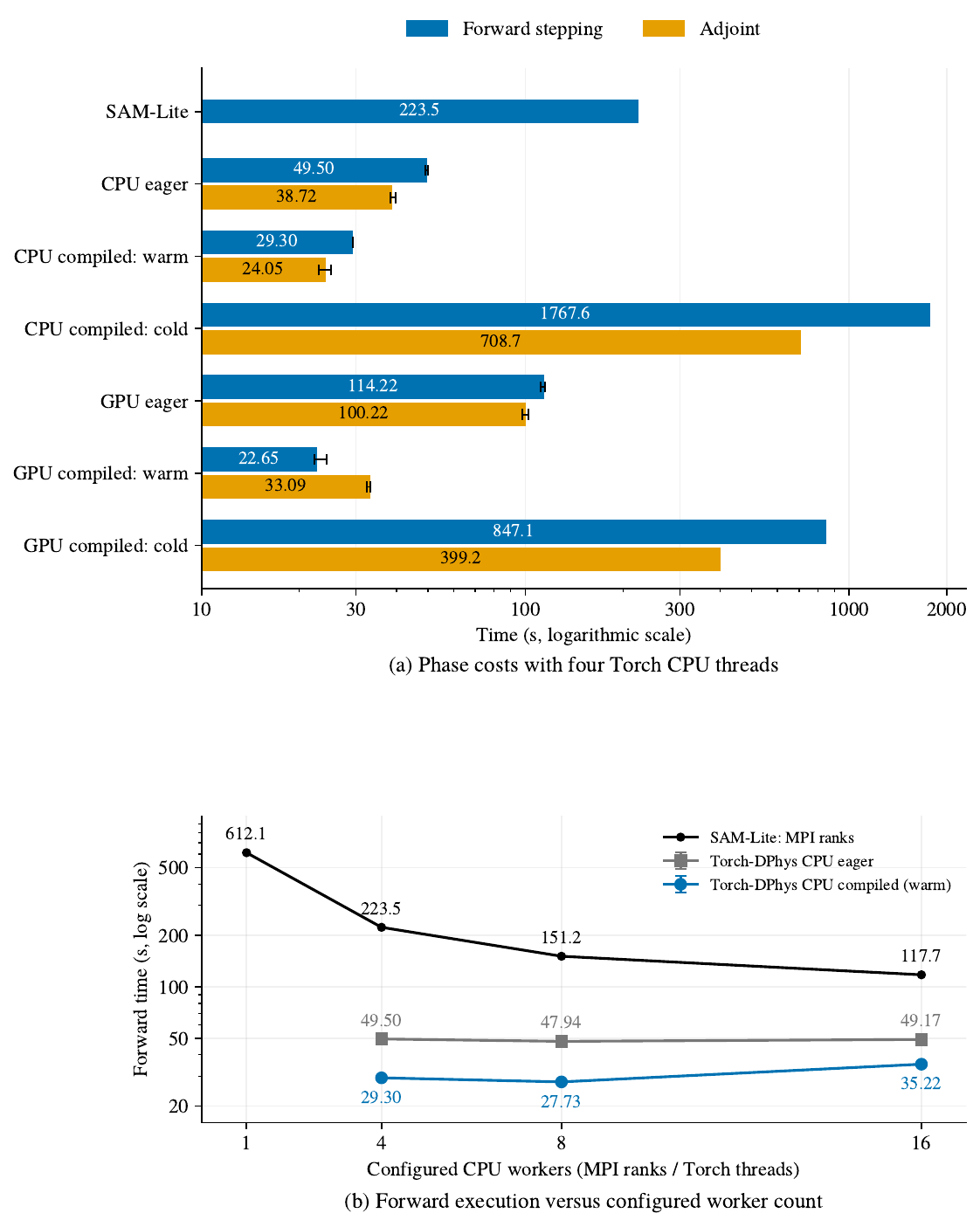}
 \caption{Wall-clock execution costs on the 200-step MSFR benchmark. (a) Forward and discrete-adjoint runtimes across 4-thread CPU and GPU configurations in Torch-DPhys (showing warm medians, min--max ranges, and cold first-call totals) alongside 4-rank MPI SAM-Lite. (b) Forward runtime versus configured CPU threads (Torch-DPhys) or MPI ranks (SAM-Lite).}
 \label{fig:msfr_200_cost}
\end{figure}

Figure~\ref{fig:msfr_200_cost}(a) compares median wall-clock times for the 200-step forward transient and 200-step discrete adjoint using four CPU threads against GPU execution and four-rank MPI SAM-Lite. On the CPU, the compiled configuration reduces median forward runtime from $49.50$~s to $29.30$~s ($1.69\times$ speedup). Both eager and compiled GPU execution use CUDA. The optimized GPU configuration combines compiled numerical kernels with CUDA graph replay, reducing median forward runtime from $114.22$~s to $22.65$~s. This improvement reflects the combined execution optimizations. For the discrete adjoint, compilation combined with auxiliary-variable reuse reduces median runtime from $38.72$~s to $24.05$~s on the CPU and from $100.22$~s to $33.09$~s on the GPU. While GPU execution achieves the fastest warm forward time ($22.65$~s), four-thread CPU execution achieves the fastest warm adjoint time among the configurations in panel (a) ($24.05$~s). First-execution forward totals, including compilation and execution, are $847.1$~s on the GPU and $1767.6$~s on the CPU. Separate compiler-time measurements for the backward-Euler and BDF2 graph specializations appear in Appendix~\ref{app:compile_cost}.

Compiled warm measurements pass background-load screening. Earlier eager medians include some load-flagged samples, whereas SAM-Lite timings are single observations with uncertain load accounting. Timing boundaries and sessions differ between codes, and GPU timings remain provisional under WSL2. Warm compiled adjoints use auxiliary reuse, while the earlier eager and cold compiled adjoints were measured without reuse.

Figure~\ref{fig:msfr_200_cost}(b) shows forward wall-clock scaling across CPU worker counts (4, 8, and 16 threads for Torch-DPhys; 1, 4, 8, and 16 MPI ranks for SAM-Lite). SAM-Lite forward runtime decreases from $612.1$~s (1 rank) to $223.5$~s (4 ranks), $151.2$~s (8 ranks), and $117.7$~s (16 ranks). Torch-DPhys compiled CPU forward runtime reaches a minimum of $27.73$~s at 8 threads ($29.30$~s at 4 threads and $35.22$~s at 16 threads). Note that Torch-DPhys uses exact AD Jacobians with direct sparse LU solves (266 total linear solves), whereas SAM-Lite uses PJFNK with finite-difference matrix-vector products (46,659 residual evaluations in the 1-rank run), so the two curves reflect both algorithmic and parallel-implementation differences.

\subsection{Component-Level Cost Breakdown}
\label{sec:msfr_component_cost}

\begin{figure}[!htbp]
 \centering
 \includegraphics[width=\linewidth,height=0.76\textheight,keepaspectratio]{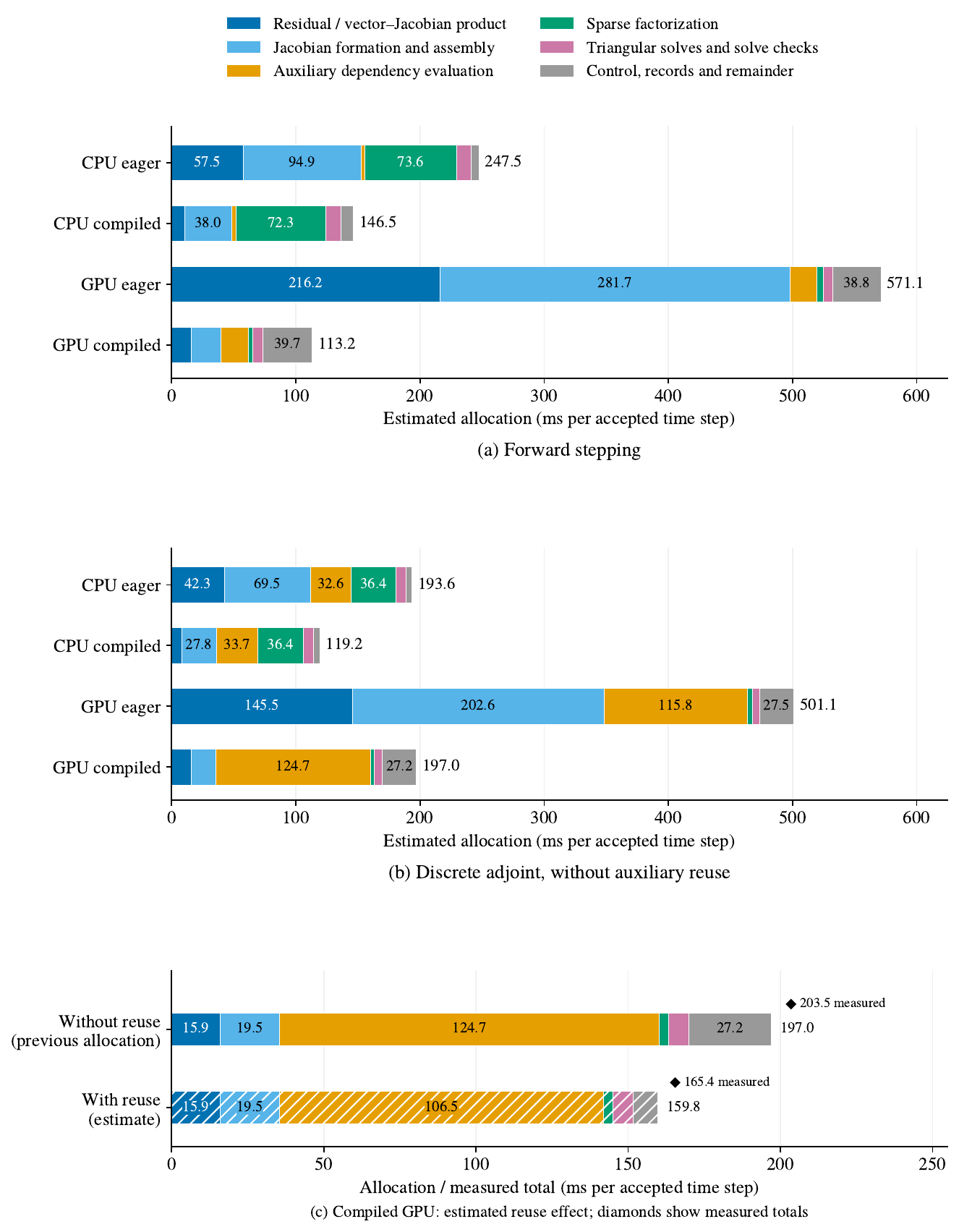}
 \caption{Per-step component runtime breakdown (in ms per accepted step) for (a) forward stepping and (b) the discrete adjoint prior to auxiliary reuse. (c) Compiled GPU adjoint breakdown before and after enabling auxiliary-variable reuse (hatched), with diamonds indicating the independently measured uninstrumented step totals.}
 \label{fig:msfr_component_cost}
\end{figure}

To explain the performance trends observed in Figure~\ref{fig:msfr_200_cost}, Figure~\ref{fig:msfr_component_cost} allocates the per-step forward and adjoint execution times to constituent numerical operations. In panels (a,b), synchronized profiling shares are scaled to the corresponding uninstrumented timing medians. These are estimated allocations of measured totals, rather than separately measured uninstrumented component costs. Panel (c) reconstructs the post-reuse component breakdown by subtracting separately profiled costs of removed auxiliary work while holding the remaining allocations fixed. Its diamond markers show independently measured totals from the later interleaved reuse comparison.

In the compiled CPU forward pass (Figure~\ref{fig:msfr_component_cost}a), single-threaded SuperLU sparse factorization accounts for $72.3$~ms per step ($49\%$ of total forward time), compared with only $3.8$~ms per step for cuDSS factorization on the GPU ($11.9$~ms including triangular solves and checks). This serial factorization limits the work that can benefit from additional Torch threads, but the baseline profile does not establish the cause of the slowdown at 16 threads on this 1,399-node mesh.

Conversely, in the compiled GPU adjoint pass (Figure~\ref{fig:msfr_component_cost}b,c), the compiled VJP kernel itself requires only $15.9$~ms per step (down from $145.5$~ms in eager GPU execution), and the transposed sparse solve requires $10.2$~ms per step. However, the GPU adjoint is bottlenecked by an uncompiled primal residual evaluation used for dynamic auxiliary-variable dependency discovery. Reusing auxiliary evaluations across the adjoint step removes two of the three redundant evaluations, reducing the GPU adjoint time from $40.70$~s ($203.5$~ms/step) to $33.09$~s ($165.4$~ms/step, an $18.7\%$ reduction), as shown in Figure~\ref{fig:msfr_component_cost}(c). The separately profiled uncompiled primal residual used for dependency discovery costs approximately $91.5$~ms per step, equivalent to about $57\%$ of the reconstructed post-reuse allocation. Reducing this work while retaining the dependency checks is a target for further GPU adjoint acceleration.

\section{Applications Enabled by Differentiability}
\label{sec:applications}

\subsection{Inferring Physical Parameters from Transient Observations}
\label{sec:parameter_identification}

We first demonstrate PDE-constrained inverse parameter estimation on the closed primary thermal-hydraulic loop (\texttt{04\_loop.i} variant verified in Figure~\ref{fig:forward04}a,b). The objective is to simultaneously recover the constant pump head $H$ and core heating power $Q$ from sparse transient measurements of core outlet mass flow rate and outlet fluid temperature recorded at $N_t = 12$ observation times over the $500$~s transient:
\begin{equation}
 t_k\in\{1,3,7,12,22,47,97,147,197,297,397,500\}~\mathrm{s}.
 \label{eq:inverse_observation_times}
\end{equation}
We minimize the scaled least-squares objective
\begin{equation}
 J(\boldsymbol\theta)=\frac{1}{2N_tN_o}\sum_{k=1}^{N_t}\sum_{j=1}^{N_o}
 \left[\frac{o_j(t_k;\boldsymbol\theta)-o_{j,k}^{\mathrm{obs}}}{s_j}\right]^2,
 \qquad N_t=12,\quad N_o=2,
 \label{eq:inverse_loss}
\end{equation}
where $s_1 = 0.01$~kg/s and $s_2 = 100$~K normalize the flow and temperature residuals to comparable magnitudes. Optimization is performed in dimensionless logarithmic coordinates $z_i = \log(\theta_i/\theta_i^{\mathrm{nom}})$ using the limited-memory Broyden--Fletcher--Goldfarb--Shanno algorithm with bound constraints (L-BFGS-B)~\cite{byrd1995,scipy116lbfgsb} with box bounds $z_i \in [\log(0.5), \log(2.0)]$ (see Appendix~\ref{app:inverse_optimizer}).

We compare both differentiation routes from Section~\ref{sec:differentiation_routes}: the discrete residual adjoint of the converged backward-Euler steps, and unrolled reverse-mode AD through $K=3$ Newton iterations per time step (using step-wise checkpoint recomputation to bound memory). Central finite differences at the initial guesses confirm that both routes match their respective forward evaluations to within $1.1\times 10^{-6}$ relative error (Appendix~\ref{app:inverse_initial_checks}).

\begin{figure}[ht!]
 \centering
 \includegraphics[width=\linewidth]{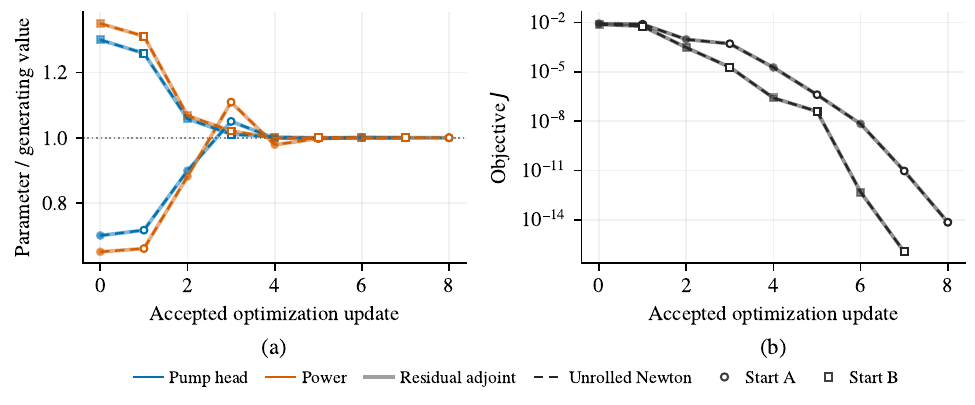}
 \caption{Optimization trajectories for same-solver parameter recovery in the \texttt{04\_loop.i} thermal-hydraulic loop: (a) normalized parameter estimates $\theta_i/\theta_i^{\mathrm{nom}}$ and (b) objective value $J$ versus L-BFGS-B update. Solid and dashed curves denote the discrete residual adjoint and unrolled Newton differentiation, respectively.}
 \label{fig:inverse}
\end{figure}

Table~\ref{tab:inverse} and Figure~\ref{fig:inverse} report parameter recovery from two initial guesses—Start A ($-30\%$ in $H$, $-35\%$ in $Q$) and Start B ($+30\%$ in $H$, $+35\%$ in $Q$)—using synthetic observations generated by the same solver ($H^{\mathrm{nom}}=0.0945$~m, $Q^{\mathrm{nom}}=2000$~W) as well as synthetic observations generated by MOOSE. On same-solver data, both the residual adjoint and unrolled AD converge in 7--8 L-BFGS-B updates along nearly indistinguishable optimization trajectories (differing by less than $1.3\times 10^{-8}$ at matched iterations) and recover the exact generating parameters. When fitting the MOOSE-generated observations from Start A, the residual-adjoint optimization converges in 8 updates to $H = 0.094516$~m ($+0.017\%$ error) and $Q = 1998.87$~W ($-0.057\%$ error), where the small residual bias reflects the minor forward discretization differences between Torch-DPhys and MOOSE noted in Section~\ref{sec:verification}.

\begin{table}[ht!]
 \centering
 \caption{Parameter identification results for the closed thermal-hydraulic loop ($H^{\mathrm{nom}}=0.0945$~m, $Q^{\mathrm{nom}}=2000$~W).}
 \label{tab:inverse}
 \small\setlength{\tabcolsep}{4pt}
 \begin{tabular}{lrrrr}
 \toprule
 Observation source / Initial guess & Initial $H$ (m) & Initial $Q$ (W) & L-BFGS-B updates & Recovered $(H,Q)$ (m, W)\\
 \midrule
 Same solver / Start A & 0.06615 & 1300 & 8 & (0.094500, 2000.00)\\
 Same solver / Start B & 0.12285 & 2700 & 7 & (0.094500, 2000.00)\\
 MOOSE / Start A & 0.06615 & 1300 & 8 & (0.094516, 1998.87)\\
 \bottomrule
 \end{tabular}
\end{table}

Although both differentiation routes yield identical parameter recovery on this well-converged problem, their computational costs differ substantially. Table~\ref{tab:backward_cost} compares the wall-clock backward time and peak live CUDA memory allocation for a single 122-step trajectory evaluation at Start A (uncompiled \texttt{float64} on an RTX 2000 Ada GPU; protocol in Appendix~\ref{app:measurement}). By solving a single transposed linear system per step rather than retracing and differentiating three Newton updates and Jacobian constructions per step, the residual adjoint reduces backward pass runtime by a factor of $10.0\times$ ($26.9$~s vs.\ $268$~s) and peak memory allocation by $8.1\times$ ($24.7$~MiB vs.\ $201.0$~MiB).

\begin{table}[htbp]
\centering
\caption{Backward-pass runtime (median and [min, max] over three fresh processes) and peak live GPU memory allocation for a 122-step trajectory at Start A.}
\label{tab:backward_cost}
\small
\begin{tabular}{lcc}
\toprule
Differentiation route & Backward time (s) & Peak allocation (MiB)\\
\midrule
Converged-residual discrete adjoint & 26.9 [26.1, 27.0] & 24.7\\
Unrolled Newton AD (with step recomputation) & 268 [263, 274] & 201.0\\
\bottomrule
\end{tabular}
\end{table}

\subsection{Solver-in-the-Loop Differentiable Predictive Control}
\label{sec:hybrid}

\begin{figure}[htbp]
 \centering
 \includegraphics[width=0.90\linewidth]{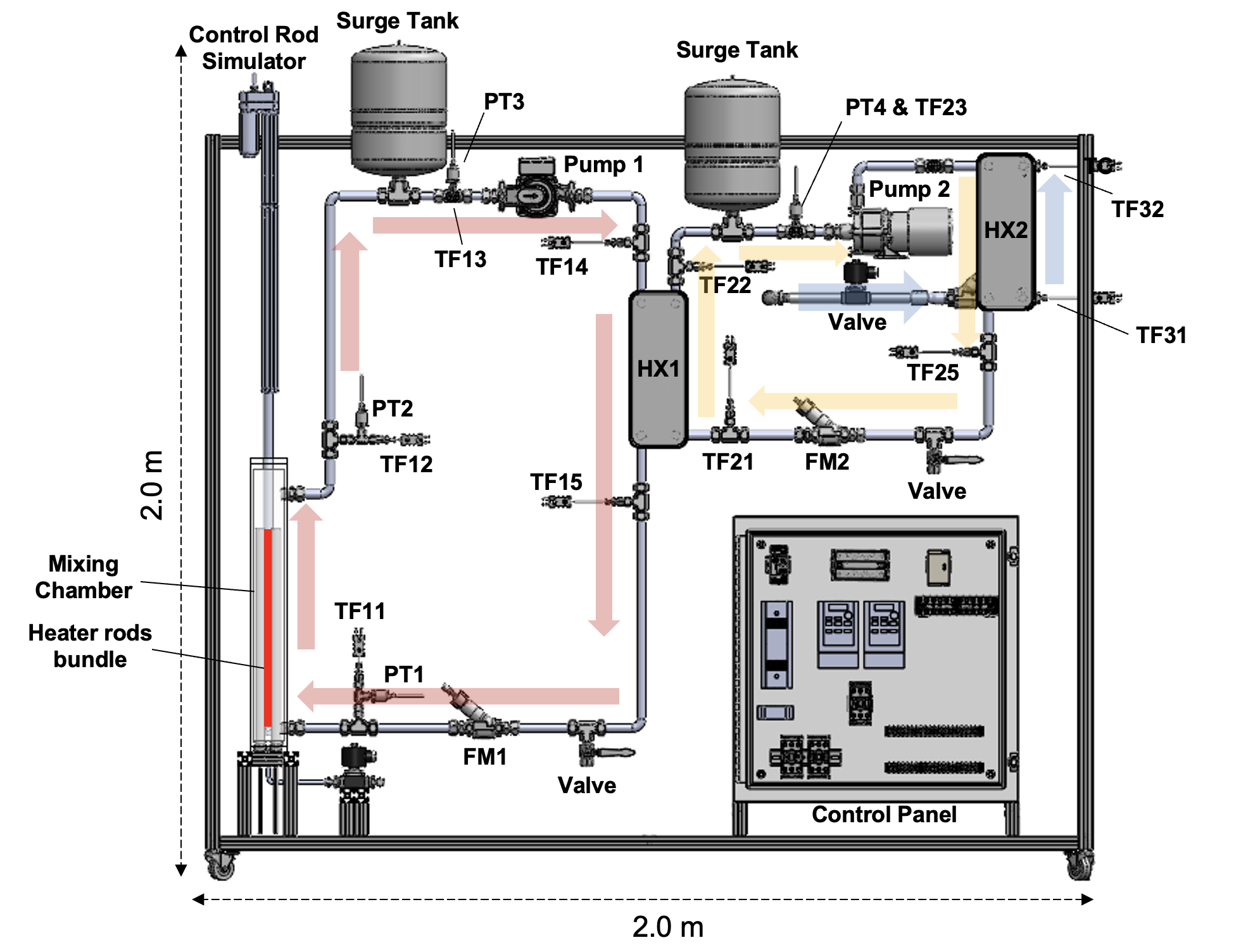}
 \caption{CAD representation of the three-loop thermal-fluid facility modeled in the differentiable predictive control demonstration (HX: heat exchanger).}
 \label{fig:experimental_facility}
\end{figure}

Beyond calibrating scalar physical parameters, end-to-end differentiability enables solver-in-the-loop training of neural network components embedded in a dynamical system. We demonstrate this capability using differentiable predictive control (DPC)~\cite{drgona2022} to train a closed-loop neural feedback controller for the three-loop electrically heated thermal-fluid test facility whose computer-aided design (CAD) representation is shown in Figure~\ref{fig:experimental_facility}. In previous work on this facility~\cite{lim2026facility,11683640}, control-oriented forecasting and policy training relied on learned Graph Neural ODE surrogates, which can introduce surrogate prediction error when the trained policy is deployed on the mechanistic plant model. Here, we eliminate the intermediate dynamics surrogate by backpropagating closed-loop control objectives directly through a PyTorch reimplementation of the facility's SAM thermal-hydraulic plant model~\cite{hu2021sam}.

\begin{figure}[htbp]
 \centering
 \resizebox{\linewidth}{!}{%
 \begin{tikzpicture}[
  font=\small,
  box/.style={draw,rounded corners=2pt,fill=black!3,align=center,text width=2.8cm,minimum height=1.0cm,inner sep=4pt},
  flow/.style={-{Stealth[length=2mm]},line width=0.6pt},
  grad/.style={flow,densely dashed}
 ]
  \node[box] (scenario) at (0,0) {Scenarios and\\demand setpoints};
  \node[box] (policy) at (3.7,0) {GRU policy\\trainable weights};
  \node[box] (plant) at (7.4,0) {Differentiable\\plant solver};
  \node[box] (loss) at (11.1,0) {Trajectory objective\\tracking and penalties};
  \draw[flow] (scenario)--(policy);
  \draw[flow] (policy)--(plant);
  \draw[flow] (plant)--(loss);
  \draw[flow] (plant.north)--(7.4,1.1)--node[above,font=\scriptsize]{simulated measurements}(3.7,1.1)--(policy.north);
  \draw[grad] (loss.south)--(11.1,-1.0)--node[below,font=\scriptsize]{objective gradients}(7.4,-1.0)--(plant.south);
  \draw[grad] (plant.south)--(7.4,-1.65)--node[below,font=\scriptsize]{backpropagate and update policy}(3.7,-1.65)--(policy.south);
  \node at (5.55,-2.35) {(a) Inner loop: solver-in-the-loop policy training via trajectory backpropagation};

  \node[box] (weights) at (0,-3.85) {Loss weights and\\temporal discount};
  \node[box] (train) at (3.7,-3.85) {Policy training\\inner loop (a)};
  \node[box] (eval) at (7.4,-3.85) {Policy\\evaluation};
  \node[box] (bo) at (11.1,-3.85) {Bayesian\\optimization};
  \draw[flow] (weights)--(train);
  \draw[flow] (train)--(eval);
  \draw[flow] (eval)--(bo);
  \draw[flow] (bo.south)--(11.1,-4.95)--node[below,font=\scriptsize]{propose next objective hyperparameters}(0,-4.95)--(weights.south);
  \node at (5.55,-5.7) {(b) Outer loop: Bayesian optimization of objective hyperparameters};
 \end{tikzpicture}}
 \caption{Two-level hybrid physics--ML training architecture for differentiable predictive control: (a) inner loop backpropagating trajectory losses through the mechanistic plant solver to update GRU policy weights, and (b) outer Bayesian optimization loop tuning objective weights and temporal discount factors.}
 \label{fig:dpc_training}
\end{figure}
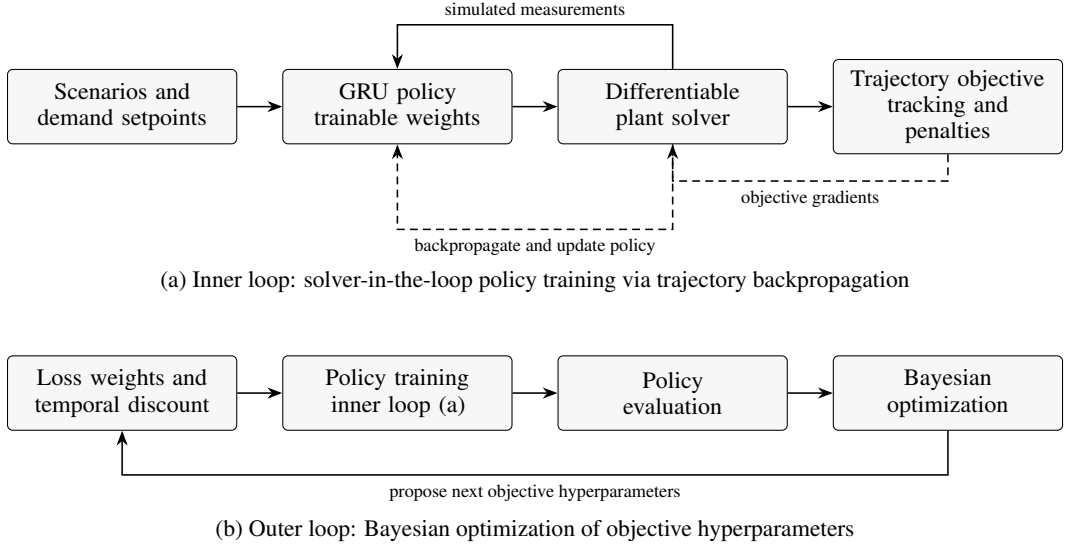

As illustrated in Figure~\ref{fig:dpc_training}(a), the neural policy is parameterized as a single-layer Gated Recurrent Unit (GRU) with 128 hidden units. At each $1$~Hz control interval, the GRU receives 12 simulated plant sensor measurements along with conditioning inputs (upcoming power demand setpoint, secondary inlet temperature, and integrated tracking error) and outputs four actuator commands regulating the core heater power, primary and intermediate pump heads, and tertiary loop mass flow rate. The differentiable plant solver advances the coupled thermal-hydraulic state over five $0.2$~s substeps per control interval across a $60$~s horizon (300 plant steps total). A composite trajectory loss penalizing thermal power tracking error, outlet temperature deviation, actuator slew rate, and heater-rod temperature exceedance above a $97\,^\circ\mathrm{C}$ soft barrier (below the $100\,^\circ\mathrm{C}$ safety limit) is then differentiated backward through all 300 plant steps to update the GRU weights via gradient descent.

To balance the competing terms in the control objective, an outer Bayesian optimization loop (Figure~\ref{fig:dpc_training}b), initialized with 20 Sobol trials followed by 20 Gaussian-process-guided trials across four independent chains (160 trained policies total, following the ensemble tuning strategy of Abulawi et al.~\cite{abulawi2025bayesian}), selects the loss weights and temporal discount factor. Because all plant rollouts and backpropagation occur offline during training, online deployment requires only a single forward pass of the 128-unit GRU at each $1$~Hz control step.

\begin{figure}[htb]
 \centering
 \includegraphics[width=\linewidth]{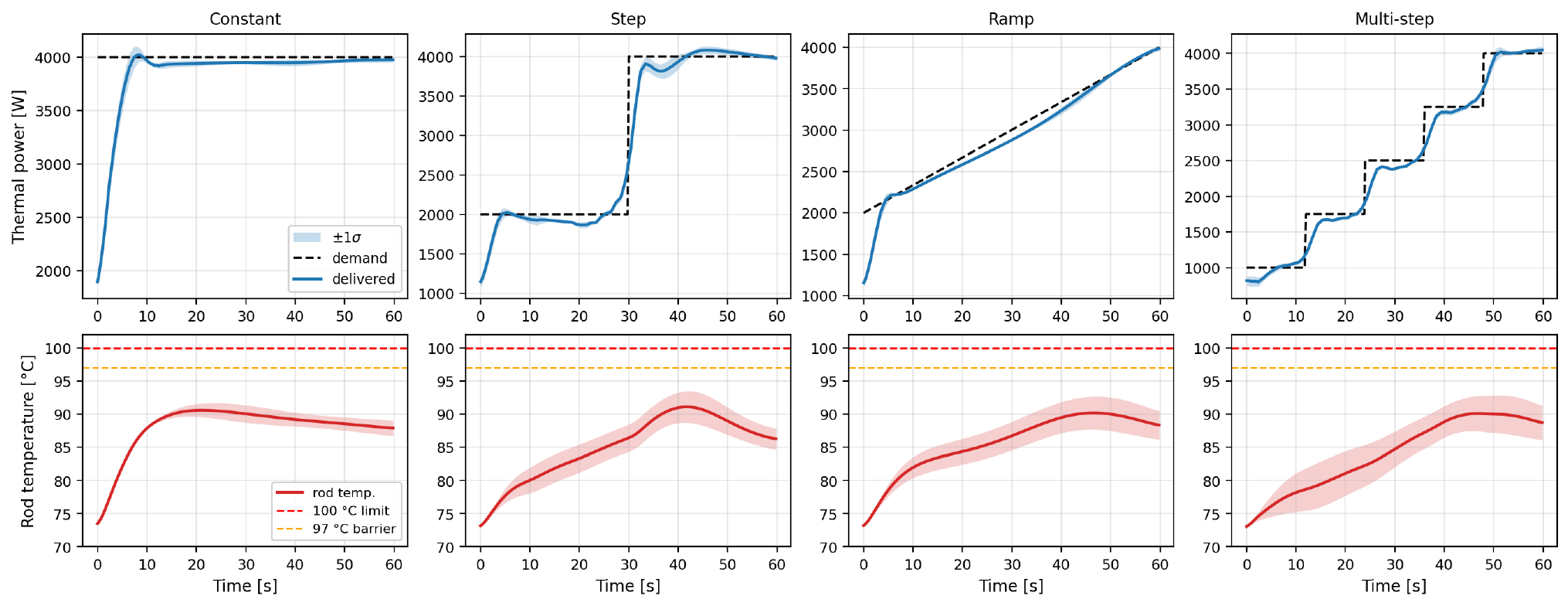}
 \caption{Closed-loop DPC tracking performance across four power-demand profiles. Top row: delivered versus demanded thermal power. Bottom row: peak heater-rod temperature relative to the $97\,^\circ\mathrm{C}$ soft penalty barrier and $100\,^\circ\mathrm{C}$ operating limit. Solid curves and shaded bands indicate the mean and $\pm 1$ standard deviation across the four selected policies.}
 \label{fig:dpc}
\end{figure}

Figure~\ref{fig:dpc} shows the closed-loop responses of the four best-performing policies (one selected from each Bayesian optimization chain) across four representative power-demand profiles. Across the four trajectories, the policies achieve full-window mean absolute power-tracking errors between $99$~W and $157$~W while constraining the maximum heater-rod surface temperature to $94.5\,^\circ\mathrm{C}$, below both the $97\,^\circ\mathrm{C}$ soft penalty barrier and the $100\,^\circ\mathrm{C}$ limit.

\subsection{Prospective Extensions to Uncertainty Quantification and Bayesian Calibration}
\label{sec:uncertainty}

Combining batched tensor execution with verified solver sensitivities also provides the mathematical building blocks for forward uncertainty propagation and Bayesian inverse calibration~\cite{smith2013uq,saltelli2008gsa,royoberkampf2011vvuq,trucano2006calibration}. Let $\mathbf q = \mathcal M(\boldsymbol\theta, \mathbf u)$ denote simulated quantities of interest as a function of uncertain model parameters $\boldsymbol\theta$ and operating inputs $\mathbf u$. Alongside Monte Carlo ensemble propagation, the exact solver Jacobian $\mathbf J_{\mathbf x} = \partial\mathcal M/\partial\mathbf x|_{\overline{\mathbf x}}$ with respect to the combined input vector $\mathbf x = [\boldsymbol\theta^T, \mathbf u^T]^T$ provides a first-order local covariance approximation:
\begin{equation}
 \boldsymbol\Sigma_{\mathbf q}\approx
 \mathbf J_{\mathbf x}\boldsymbol\Sigma_{\mathbf x}\mathbf J_{\mathbf x}^T,
 \qquad
 \mathbf J_{\mathbf x}=
 \left.\frac{\partial\mathcal M}{\partial\mathbf x}\right|_{\overline{\mathbf x}},
 \label{eq:uq_covariance}
\end{equation}
as well as derivative-based global sensitivity measures~\cite{sobolkucherenko2009dgsm}.

Similarly, the deterministic point estimation in Section~\ref{sec:parameter_identification} can be extended to full Bayesian posterior inference under the statistical observation model~\cite{kennedyohagan2001,brynjarsdottir2014discrepancy}
\begin{equation}
 \mathbf o^{\mathrm{obs}}=\mathcal H\!\left[\mathcal M(\boldsymbol\theta,\mathbf u)\right]
 +\boldsymbol\delta+\boldsymbol\varepsilon,
 \label{eq:uq_observation}
\end{equation}
where $\mathcal H$ is the observation operator, $\boldsymbol\delta$ represents structural model discrepancy, and $\boldsymbol\varepsilon$ represents measurement noise. In this setting, the verified residual-adjoint gradients enable gradient-based Markov chain Monte Carlo (such as Hamiltonian Monte Carlo) or variational inference over high-dimensional closure parameters and embedded neural network weights.

\section{Discussion and Conclusions}
\label{sec:conclusion}

We have presented Torch-DPhys, a PyTorch/LibTorch-native framework that unifies selected finite-element and finite-volume physical models from MOOSE THM, SAM, and continuum mechanics with automatic differentiation and machine learning workflows. By lowering object-oriented component definitions and finite-element spaces into tensor operators, Torch-DPhys evaluates physical residuals, assembles sparse Jacobians, and propagates trajectory sensitivities within the same computational environment used to train neural networks and run gradient-based optimizers. The framework supports both unrolled differentiation through executed solver steps and discrete adjoints of converged implicit residuals, allowing users to trade off implementation simplicity against backward-pass memory and compute cost.

Forward numerical verification confirms that the implemented tensor operators reproduce the discrete behavior of the reference codes on the tested configurations. In two-dimensional plane-strain elasticity, Torch-DPhys matches archived MOOSE nodal displacements to within $2.4\times 10^{-13}$ relative $L_2$ difference. Across three one-dimensional THM channel and loop transients—incorporating HLLC convective fluxes, volume and parallel-channel junctions, conjugate heat structures, pumps, counterflow heat exchangers, and PID control—Torch-DPhys reproduces MOOSE temperature histories within $0.096$~K and mass-flow histories within $0.14\%$. On the two-dimensional axisymmetric MSFR initialization transient, Torch-DPhys and SAM-Lite independently select identical 200-step adaptive schedules and agree to within a maximum normalized field difference of $1.13\times 10^{-5}$ ($1.94\times 10^{-5}$~K in temperature and $2.85\times 10^{-2}$~Pa in pressure). Transient sensitivity verification confirms that the BDF2 discrete adjoint matches tangent-linear sensitivities to $7.37\times 10^{-10}$ across 39 parameter directions on 40-step windows and to $4.2\times 10^{-13}$ across four directions over the full 200-step trajectory.

On the 200-step MSFR benchmark, combining sparse direct linear algebra (SuperLU on CPU and cuDSS on GPU) with \texttt{torch.compile} reduces median forward execution time from $49.50$~s to $29.30$~s on a four-thread CPU configuration and from $114.22$~s to $22.65$~s on an NVIDIA RTX 2000 Ada GPU. Reusing auxiliary-variable evaluations across the adjoint dependency checks lowers the compiled GPU adjoint runtime by $18.7\%$ (to $33.09$~s), compared with $24.05$~s on the four-thread CPU. Component-level profiling indicates that CPU forward runtime is primarily bounded by single-threaded sparse LU factorization ($49\%$ of step cost), whereas GPU adjoint runtime is bounded by uncompiled host/device dependency discovery (approximately $57\%$ of the reconstructed post-reuse allocation).

The two application studies illustrate how these verified gradients support inverse problems and hybrid physics--ML. In the 1D thermal-hydraulic loop, L-BFGS-B optimization driven by either unrolled AD or the discrete residual adjoint recovers nominal pump head and heating power from sparse transient observations, achieving relative errors below $0.06\%$ when fitting synthetic MOOSE data while requiring $10\times$ less backward time and $8\times$ less peak memory with the residual adjoint than with checkpointed unrolled differentiation. In the three-loop facility demonstration, differentiating a 300-step closed-loop trajectory objective directly through the SAM-derived plant solver trains a GRU feedback policy that tracks multi-step power demands while maintaining peak heater rod temperatures below the $100\,^\circ\mathrm{C}$ limit, avoiding the surrogate mismatch incurred when training controllers on standalone neural ODE emulators.

Several scope boundaries qualify the reported results and define directions for future work. First, while close agreement between Torch-DPhys and MOOSE or SAM-Lite verifies that the tensor implementations reproduce the reference discrete formulations at the tested tolerances, cross-code agreement does not by itself establish spatial or temporal continuum convergence. In the axisymmetric MSFR study, because the standard SUPG/PSPG stabilization coefficients $\tau_c$, $\tau_m$, and $\tau_e$ depend explicitly on $\Delta t$, refining the time step alters the stationary discrete spatial operator, shifting the endpoint inlet pressure by $5.8\%$ and retaining a $1.65\text{--}1.90\%$ velocity-field mass-flux imbalance on the coarse mesh. Systematic spatial mesh refinement and time-step-independent stabilization formulations are therefore needed to establish continuum accuracy. In addition, the MSFR transient adjoint differentiates trajectories conditional on a fixed time-step schedule and recorded retry history; it does not differentiate through adaptive step-size selection or discrete branch switches, nor does it currently account for parameter dependence in the initial state.

From an architectural standpoint, the structural mechanics, one-dimensional THM, reduced SAM control plant, and two-dimensional axisymmetric MSFR modules currently operate with separate data structures and solver interfaces. Unifying these interfaces and exposing the MSFR discrete adjoint as an end-to-end PyTorch autograd function will extend solver-in-the-loop training from control policies to embedded neural turbulence closures—building on offline SAM-ML workflows~\cite{liu2022samml,liu2024msrsamml}—as well as coupled fluid--structure interaction and MOOSE MultiApp integration via LibTorch. Eliminating the remaining uncompiled dependency-probing overhead in the GPU adjoint and reusing symbolic sparse factorizations will further reduce turnaround times for multi-step gradient evaluations. Finally, because the parameter-identification and DPC demonstrations reported here use noiseless synthetic observations and simulated plant rollouts, future work will evaluate these workflows on noisy measurements from instrumented test facilities and incorporate the ensemble uncertainty propagation and Bayesian calibration methods outlined in Section~\ref{sec:uncertainty}.

% \section*{Code and Data Availability}

% The manuscript source bundle contains the LaTeX source files, bibliography, and figure assets; the Torch-DPhys solver source code, MOOSE and SAM-Lite reference input decks, meshes, and simulation archives are maintained in internal development repositories. SAM-derived reference materials are subject to their respective software distribution terms.

\begin{ack}
This research is supported by the U.S. Department of Energy (DOE) Office of Nuclear Energy's Nuclear Energy University Program (NEUP) under Award Number DE-NE0009613.
\end{ack}

\bibliographystyle{unsrtnat}
\bibliography{references}
\clearpage

\appendix
\section{Differentiation, Optimization, and Measurement Details}
\label{app:differentiation}

While PyTorch's autograd engine provides a unified interface for propagating derivatives across computational stages, the mathematical meaning of the resulting gradient depends on whether differentiation is applied to the finite-step numerical solver or to the implicit discrete residual equations. Below we detail both formulations as implemented in Section~\ref{sec:differentiation_routes} and evaluated in Section~\ref{sec:parameter_identification}.

\subsection{Differentiating a Finite Numerical Calculation}

Let $\mathcal A_n^{(K)}$ denote the numerical update operator that advances the state at time step $n$ using $K$ nonlinear solver iterations, including its state initialization, Jacobian evaluations, and linear system solves:
\begin{equation}
 \widetilde{\mathbf y}_n=
 \mathcal A_n^{(K)}(\widetilde{\mathbf y}_{n-1},\boldsymbol\theta).
 \label{eq:algorithm_map}
\end{equation}
Applying reverse-mode AD to Eq.~\eqref{eq:algorithm_map} propagates sensitivities through the exact sequence of arithmetic and linear-algebra operations executed during those $K$ updates. Because the initial guess at step $n$ is seeded from $\widetilde{\mathbf y}_{n-1}$, parameter sensitivities propagate through both the initial guess and the $K$ Newton updates, which in turn requires second-derivative information when differentiating through the AD-constructed Jacobian. When $K$ is small and the step has not fully converged, $\partial\widetilde{\mathbf y}_n/\partial\boldsymbol\theta$ gives the exact derivative of the $K$-step map $\mathcal A_n^{(K)}$ rather than the implicit sensitivity of the root $\mathbf R_n = \mathbf 0$.

In the parameter-identification study (Section~\ref{sec:parameter_identification}), the unrolled route executes $K=3$ Newton updates per time step. To avoid retaining the full Newton-iteration graph across all 122 time steps simultaneously in GPU memory, the implementation checkpoints the accepted step states and recomputes the $K=3$ forward Newton updates locally within each step during the backward pass.

\subsection{Differentiating the Implicit Discrete Equations}

For the residual-adjoint route, write the full transient system across all prescribed time steps as $\mathbf R(\mathbf Y,\boldsymbol\theta)=\mathbf0$, where $\mathbf Y$ stacks the converged states across the time grid. For a scalar objective $J(\mathbf Y,\boldsymbol\theta)$, assuming local differentiability and an invertible block-lower-triangular trajectory Jacobian $\mathbf R_{\mathbf Y} = \partial\mathbf R/\partial\mathbf Y$, implicit differentiation yields
\begin{equation}
 \mathbf R_{\mathbf Y}^{T}\boldsymbol\lambda
 =\nabla_{\mathbf Y}J,
 \qquad
 \nabla_{\boldsymbol\theta}J_{\mathrm{total}}
 =\partial_{\boldsymbol\theta}J
 -\mathbf R_{\boldsymbol\theta}^{T}\boldsymbol\lambda.
 \label{eq:trajectory_adjoint}
\end{equation}
Because $\mathbf R_{\mathbf Y}$ is block lower-bidiagonal (for backward Euler) or block lower-tridiagonal (for BDF2) in time, $\mathbf R_{\mathbf Y}^{T}\boldsymbol\lambda = \nabla_{\mathbf Y}J$ is solved by backward substitution one time step at a time (Eq.~\eqref{eq:adjoint} and Eq.~\eqref{eq:msfr_bdf2_adjoint}).

Evaluating Eq.~\eqref{eq:trajectory_adjoint} depends only on the converged trajectory $\mathbf Y$ and does not retrace the forward Newton path or line-search history. Consequently, only one transposed linear solve is required per accepted time step, and no second-order derivatives of the residual are needed. However, any nonzero forward residual norm $\|\mathbf R_n\|$ or inexactness in solving the transposed linear system introduces an approximation error relative to the exact implicit derivative.

\subsection{Scope of the Parameter-Recovery Comparison}

In Section~\ref{sec:parameter_identification}, both differentiation routes are evaluated on the same 122-step time grid, physical initial state, observation times, and box bounds. The synthetic same-solver observations are generated once from a nominal converged trajectory ($H^{\mathrm{nom}}=0.0945$~m, $Q^{\mathrm{nom}}=2000$~W).

Gradient accuracy is verified in the dimensionless logarithmic coordinates used by the optimizer, $z_i = \log(\theta_i/\theta_i^{\mathrm{nom}})$, where $\partial J/\partial z_i = \theta_i\,\partial J/\partial\theta_i$. Each differentiation route is checked against central finite differences of its own corresponding forward evaluation ($K=3$ Newton updates for the unrolled route, and tolerance-converged solves for the residual-adjoint route). Because $K=3$ Newton iterations reduce the step residuals nearly to machine precision on this 1D loop problem, the two routes produce nearly identical gradients and optimization paths in Figure~\ref{fig:inverse}.

\subsection{Residual-Adjoint Gradient Checks Away from the Optimum}
\label{app:inverse_initial_checks}

Table~\ref{tab:inverse_initial_gradients} summarizes the central finite-difference verification of the residual-adjoint gradients at Initial Guess A ($z_H = \log(0.7)$, $z_Q = \log(0.65)$) for both the same-solver and MOOSE observation datasets. For each coordinate $i \in \{H, Q\}$ and step size $h \in \{10^{-2}, 10^{-3}, 10^{-4}\}$, fresh forward trajectories are solved at $\mathbf z \pm h\mathbf e_i$, yielding central finite-difference (FD) derivatives compared via the relative error $|g_i - g_i^{\mathrm{FD}}|/|g_i^{\mathrm{FD}}|$.

\begin{table}[htbp]
 \centering
 \caption{Comparison of residual-adjoint logarithmic derivatives against central finite differences at Initial Guess A ($z_H=\log(0.7)$, $z_Q=\log(0.65)$) for perturbation step sizes $h \in \{10^{-2}, 10^{-3}, 10^{-4}\}$.}
 \label{tab:inverse_initial_gradients}
 \small
 \begin{tabular}{llrrrr}
 \toprule
 Observations & Parameter & Adjoint derivative & $h=10^{-2}$ & $h=10^{-3}$ & $h=10^{-4}$\\
 \midrule
 Same solver & Pump head & $-0.02324309$ & $7.63\times10^{-5}$ & $6.13\times10^{-7}$ & $1.93\times10^{-7}$\\
 Same solver & Heating power & $-0.01525549$ & $9.90\times10^{-5}$ & $1.01\times10^{-6}$ & $1.20\times10^{-6}$\\
 MOOSE & Pump head & $-0.02326628$ & $7.62\times10^{-5}$ & $6.12\times10^{-7}$ & $1.93\times10^{-7}$\\
 MOOSE & Heating power & $-0.01522336$ & $9.92\times10^{-5}$ & $1.02\times10^{-6}$ & $1.20\times10^{-6}$\\
 \bottomrule
 \end{tabular}
\end{table}

At $h = 10^{-3}$, all four derivatives agree with central finite differences to within $1.02\times 10^{-6}$; at $h = 10^{-4}$, finite-precision cancellation in the objective differences begins to limit further reduction of the heating-power discrepancy. At Initial Guess B ($z_H = \log(1.3)$, $z_Q = \log(1.35)$), the residual-adjoint logarithmic gradient $(0.03225393, 0.02902049)$ matches the finite-difference-verified unrolled AD gradient to within a relative discrepancy of $3.9\times 10^{-9}$.

\subsection{Parameter-Identification Optimizer Settings}
\label{app:inverse_optimizer}

The L-BFGS-B optimizer~\cite{byrd1995} (SciPy 1.16.1~\cite{scipy116lbfgsb}) updates the log-parameters $z_i \in [\log(0.5), \log(2.0)]$ with a maximum of 30 iterations and up to 15 line-search trials per iteration. Optimization terminates when either the projected gradient norm or relative objective reduction satisfies
\begin{equation}
 \|\mathbf g_{\mathrm{proj}}\|_\infty\le10^{-7}
 \quad\text{or}\quad
 \frac{J_k-J_{k+1}}{\max(|J_k|,|J_{k+1}|,1)}\le10^{-13}.
 \label{eq:inverse_optimizer_stopping}
\end{equation}
All optimization runs in Table~\ref{tab:inverse} converged via the projected-gradient criterion $\|\mathbf g_{\mathrm{proj}}\|_\infty \le 10^{-7}$, requiring 10 total objective/gradient evaluations for Start A (8 accepted updates) and 8 evaluations for Start B (7 accepted updates).

\subsection{Protocol for Differentiation-Route Cost Measurements}
\label{app:measurement}

The backward runtime and memory comparisons in Table~\ref{tab:backward_cost} were measured over the 122-step trajectory at Initial Guess A in \texttt{float64} precision on an NVIDIA RTX 2000 Ada GPU (driver 595.95, PyTorch 2.8.0+cu128) hosted by an Intel Core Ultra 7 265K CPU (pinned to logical CPU 0 with a single intra-op and inter-op thread, compilation disabled). Following a three-step warm-up, three fresh-process repetitions were executed in alternating order for each differentiation route.

Peak memory was recorded via PyTorch's CUDA memory allocator after resetting peak statistics immediately prior to the backward sweep, and therefore includes the retained 122-step forward trajectory tensors ($24.7$~MiB for the residual adjoint versus $201.0$~MiB for unrolled AD with step-wise recomputation). Wall-clock backward timings were bracketed by \texttt{torch.cuda.synchronize()} calls excluding setup and trajectory file export.

\section{Additional Numerical Verification}
\label{app:sam_diagnostics}

This appendix provides detailed residual, nonlinear convergence, time-step refinement, sparse linear algebra, and transient-adjoint diagnostics supporting the MSFR verification in Sections~\ref{sec:sam_residual}--\ref{sec:msfr_stage1_benchmark}.

\subsection{Residual, Derivative, and Recorded-Field Checks}

At eight archived SAM-Lite reference states, the assembled Torch-DPhys residual vector matches SAM-Lite bit-for-bit at 5,592 of 5,596 entries in four states and at all 5,596 entries in the remaining four states; the four-entry differences are 1~ULP discrepancies at the inlet Dirichlet nodes resulting from floating-point evaluation order in the inlet velocity profile.

Local automatic derivatives of the MSFR residual with respect to the current state $\mathbf y_k$, lagged states $(\mathbf y_{k-1}, \mathbf y_{k-2})$, and 34 registered parameter blocks were verified against central finite differences across 360 sampled active state-Jacobian entries, all constrained Dirichlet identity rows, and 1,080 scalar history-derivative entries (including across a rejected step retry).

In the 200-step transient comparison (Figure~\ref{fig:msfr_fields}d), the normalized field difference plateau from step 186 to step 200 occurs because the initial residual at those near-steady steps already satisfies the $10^{-6}$ stopping tolerance, so both SAM-Lite and Torch-DPhys accept steps 187--200 with zero Newton updates.

\subsection{Nonlinear Solution and Time Stepping}

\paragraph{Reference nonlinear configuration and matched-step convergence.}
In the reference SAM-Lite configuration, each implicit time step is solved via PETSc PJFNK using matrix-free finite-difference Jacobian--vector products, a hand-coded preconditioning matrix factored via complete LU decomposition with diagonal shifting, restarted generalized minimal residual (GMRES) iteration with restart length 100 (up to 200 linear iterations), and cubic backtracking line search. Torch-DPhys instead assembles the exact residual Jacobian via AD and applies a direct LU solve with the same cubic backtracking line search.

\begin{figure}[htbp]
 \centering
 \includegraphics[width=\linewidth]{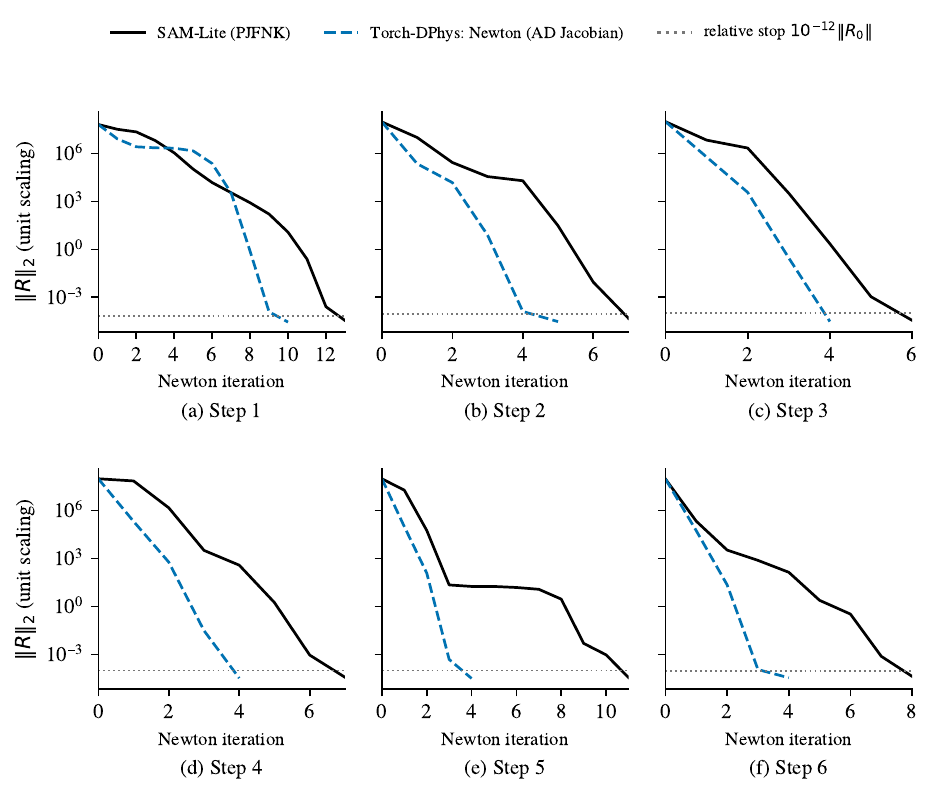}
 \caption{Nonlinear residual convergence histories on six matched single-step MSFR problems comparing SAM-Lite PJFNK against Torch-DPhys Newton with exact AD Jacobians and direct LU solves: (a) backward-Euler startup step and (b--f) BDF2 continuation steps.}
 \label{fig:msfr_newton}
\end{figure}

Figure~\ref{fig:msfr_newton} compares the nonlinear residual convergence histories of SAM-Lite (PJFNK) and Torch-DPhys (exact AD Newton) across six matched single-step problems initialized from archived SAM-Lite states under unit residual scaling. Across the six steps, SAM-Lite requires 13, 7, 6, 7, 11, and 8 nonlinear iterations to reach $10^{-12}\|\mathbf R_0\|_2$, whereas Torch-DPhys converges in 10, 5, 4, 4, 4, and 4 iterations.

\paragraph{Preconditioner Jacobian and single-correction diagnostics.}
Comparing SAM-Lite's hand-coded preconditioning matrix against a finite-difference Jacobian at $\Delta t = 0.01$~s reveals a relative Frobenius-norm difference of approximately $1.7\%$ (arising from omitted derivative contributions in the hand-coded stabilization/turbulence terms). While effective as a PJFNK preconditioner, using the hand-coded matrix directly as a true Newton Jacobian without line search diverges on the first step, whereas the exact AD Jacobian in Torch-DPhys converges quadratically. Furthermore, starting from four converged SAM-Lite states, a single exact Newton correction in Torch-DPhys has a maximum normalized magnitude of $1.3\times 10^{-14}$, confirming consistency of the discrete roots.

\paragraph{Adaptive time-stepping rules and retry lifecycle.}
The iteration-based adaptive time stepper increases $\Delta t$ by a factor of $1.1$ when a step converges in fewer than 5 nonlinear iterations (and fewer than 750 linear iterations), decreases $\Delta t$ by a factor of $0.8$ when nonlinear iterations exceed 9 (or linear iterations exceed 1,350), and cuts back by a factor of $0.8$ to retry upon a failed nonlinear solve. Following a rejected attempt, the solver restores the last accepted state and auxiliary fields; to match the MOOSE execution lifecycle, the subsequent retry retains the failed attempt's step size as \texttt{dt\_old} in the variable-step BDF2 coefficients. Standalone driver verification tests confirmed exact reproduction of a 65-attempt sequence containing four step rejections and boundary clips.

\paragraph{Stopping criteria and tight-tolerance step-rejection replay.}
Both solvers evaluate nonlinear convergence using the separate absolute and relative criteria
\begin{equation}
 \|\mathbf R\|_2<a_{\mathrm{tol}}
 \quad\text{or}\quad
 \|\mathbf R\|_2\le r_{\mathrm{tol}}\|\mathbf R_0\|_2,
 \label{eq:msfr_stop}
\end{equation}
where $\|\mathbf R\|_2$ is the Euclidean norm of the scaled residual vector. In the baseline input deck, $a_{\mathrm{tol}} = r_{\mathrm{tol}} = 10^{-6}$, with velocity and pressure rows scaled by $0.1$ and energy rows scaled by $10^{-3}$.

\begin{figure}[htbp]
 \centering
 \includegraphics[width=\linewidth]{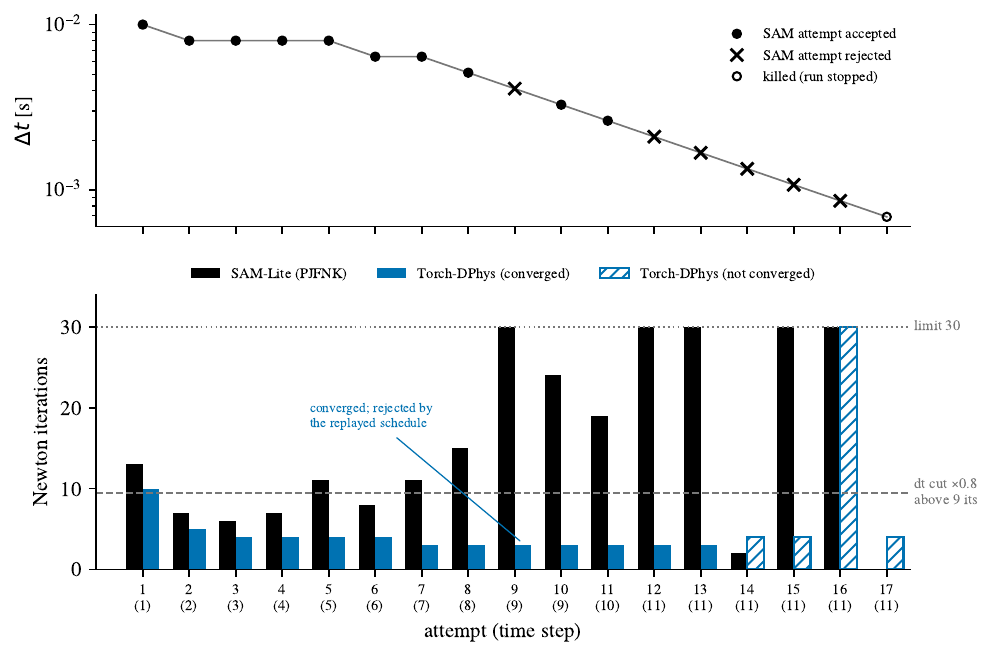}
 \caption{Attempt sequence and nonlinear iteration counts during the tight-tolerance ($r_{\mathrm{tol}}=10^{-12}$, unit scaling) MSFR replay. Top: step sizes and acceptance outcomes. Bottom: nonlinear iteration counts for SAM-Lite and Torch-DPhys.}
 \label{fig:msfr_attempts}
\end{figure}

In a separate stress test using unit row scaling, $r_{\mathrm{tol}} = 10^{-12}$, $a_{\mathrm{tol}} = 10^{-4}$, and a maximum of 30 nonlinear iterations (Figure~\ref{fig:msfr_attempts}), SAM-Lite completed 10 accepted steps ($t = 0.0658$~s) with a step rejection at attempt 9 before stalling in repeated cutbacks at step 11. Because the exact AD Newton solver in Torch-DPhys converges in fewer iterations (10, 5, 4, 4, 4, 4, 3, 3, 3, and 3 iterations on the 10 accepted steps), Torch-DPhys converged on attempt 9 where SAM-Lite rejected; to test retry history handling in the adjoint, the replay driver explicitly imposed the SAM-Lite rejection at attempt 9.

\subsection{Baseline Adaptive Schedule Settings}

In the 200-step baseline comparison (Section~\ref{sec:sam_residual}), both SAM-Lite and Torch-DPhys start from $\Delta t_1 = 0.01$~s with step bounds $[10^{-7}, 0.25]$~s, target $7 \pm 2$ nonlinear iterations, and apply $a_{\mathrm{tol}} = r_{\mathrm{tol}} = 10^{-6}$ with row scaling $(0.1, 0.1, 10^{-3})$ and a 15-iteration cap. Although nonlinear iteration counts differed slightly on 59 of the 200 steps, all counts remained within the $[1, 4]$ growth band, resulting in bitwise-identical 200-step time grids up to $t = 43.7148$~s.

\subsection{Time-Step Sensitivity and Conservation}
\label{app:msfr_timestep}

Table~\ref{tab:msfr_timestep_sensitivity} summarizes the endpoint changes observed in the Torch-DPhys constant-step refinement study ($t \in [2.71, 43.7]$~s across $\Delta t \in \{0.25, 0.125, 0.0625, 0.03125\}$~s) discussed in Section~\ref{sec:msfr_stage1_benchmark}.

\begin{table}[!htbp]
 \centering
 \caption{Endpoint solution sensitivity of the production MSFR formulation when refining a constant time step from $\Delta t = 0.25$~s to $\Delta t = 0.03125$~s at $t = 43.7$~s.}
 \label{tab:msfr_timestep_sensitivity}
 \small
 \renewcommand{\arraystretch}{1.2}
 \begin{tabular}{@{}p{0.45\linewidth}rp{0.34\linewidth}@{}}
 \toprule
 Quantity & Change ($\Delta t = 0.03125$~s minus $0.25$~s) & Context\\
 \midrule
 Mean outlet temperature & $+0.46$~K & \\
 Mean inlet pressure & $+2.2$~kPa & $5.8\%$ of the inlet--outlet pressure drop\\
 Outlet mass flow from velocity field & $+36$~kg/s & Global mass-flux imbalance rises from $1.65\%$ to $1.90\%$\\
 Maximum core temperature & $+0.8$~K & \\
 \bottomrule
 \end{tabular}
\end{table}

At the $t = 43.7148$~s endpoint of the 200-step baseline run, direct quadrature integration of the normal velocity field over the inlet and outlet boundaries gives mass flow rates of $14,918.60$~kg/s and $15,165.31$~kg/s, respectively—a net imbalance of $246.71$~kg/s ($1.65\%$), matching the archived SAM-Lite boundary flux postprocessors. Because fluid density $\rho$ is constant, this discrepancy reflects the weak enforcement of continuity with PSPG stabilization on the coarse 1,296-element mesh rather than physical mass accumulation.

\subsection{Sparse Linear Algebra and Numerical Agreement}
\label{app:msfr_linear_algebra}

Each Newton update and discrete-adjoint step solves a linear system
\begin{equation}
 \mathbf M\mathbf x=\mathbf b,
 \label{eq:msfr_linear_system}
\end{equation}
where $\mathbf M \in \mathbb R^{5596\times 5596}$ is either the state Jacobian $\mathbf A_k$ or its transpose $\mathbf A_k^T$. Whereas the reference dense path computes a pivoted LU factorization $\mathbf P\mathbf M = \mathbf L\mathbf U$ followed by triangular substitutions
\begin{equation}
 \mathbf P\mathbf M=\mathbf L\mathbf U, \qquad
 \mathbf L\mathbf z=\mathbf P\mathbf b, \qquad
 \mathbf U\mathbf x=\mathbf z,
 \label{eq:msfr_dense_lu}
\end{equation}
the sparse path retains only the 182,809 structural nonzeros ($0.58\%$ density). To mitigate ill-conditioning prior to sparse factorization, each row is equilibrated by its maximum absolute entry:
\begin{equation}
 D_{ii}=\frac{1}{\max_j|M_{ij}|},
 \qquad
 \widehat{\mathbf M}=\mathbf D\mathbf M,
 \qquad
 \widehat{\mathbf b}=\mathbf D\mathbf b.
 \label{eq:msfr_row_equilibration}
\end{equation}
The equilibrated matrix is factored with row and column permutations $\mathbf P_r$ and $\mathbf P_c$:
\begin{equation}
 \mathbf P_r\widehat{\mathbf M}\mathbf P_c
 =\mathbf L_s\mathbf U_s,
 \label{eq:msfr_sparse_lu}
\end{equation}
and the initial solution $\mathbf x^{(0)}$ is polished with one step of iterative refinement using the unscaled residual:
\begin{equation}
 \begin{aligned}
  \mathbf r &= \mathbf b-\mathbf M\mathbf x^{(0)},\\
  \widehat{\mathbf M}\,\delta\mathbf x &= \mathbf D\mathbf r,\\
  \mathbf x^{(1)} &= \mathbf x^{(0)}+\delta\mathbf x.
 \end{aligned}
 \label{eq:msfr_iterative_refinement}
\end{equation}

Table~\ref{tab:msfr_sparse_accuracy} compares the sparse SuperLU solver against the dense LU reference on the 10-step retry and 40-step verification windows, showing agreement to within $1.53\times 10^{-14}$ for forward states and $1.49\times 10^{-11}$ for trajectory gradients.

\begin{table}[!htbp]
 \centering
 \caption{Maximum normalized discrepancies between the sparse SuperLU and dense LU implementations across the 10-step retry and 40-step MSFR verification windows.}
 \label{tab:msfr_sparse_accuracy}
 \small
 \renewcommand{\arraystretch}{1.2}
 \begin{tabular}{@{}p{0.32\linewidth}p{0.40\linewidth}r@{}}
 \toprule
 Quantity & Comparison scope & Maximum discrepancy\\
 \midrule
 Jacobian assembly & Tested states within each execution mode & Bitwise identical\\
 Forward states & Eager trajectories (Eq.~\eqref{eq:msfr_field_error}) & $1.53\times10^{-14}$\\
 Adjoint gradients & Evaluated on common dense trajectories & $7.14\times10^{-12}$\\
 Adjoint gradients & Sparse forward trajectory vs.\ dense reference & $1.49\times10^{-11}$\\
 \bottomrule
 \end{tabular}
 \par\smallskip
 \begin{minipage}{\linewidth}
 \small
 Blockwise gradient discrepancies are measured over all active parameter blocks $j$ via
 \begin{equation}
  e_g=\max_j
  \frac{\|\mathbf g_j^{\mathrm{sparse}}-\mathbf g_j^{\mathrm{dense}}\|_\infty}
       {\|\mathbf g_j^{\mathrm{dense}}\|_\infty},
  \label{eq:msfr_sparse_gradient_difference}
\end{equation}
 while structurally zero parameter blocks match identically.
 \end{minipage}
\end{table}

\subsubsection{Sparse-Backend Verification}
\label{app:msfr_sparse_details}

Across 417 benchmark linear systems (50 adjoint systems, 204 trajectory Newton systems, 131 matched-step Newton systems, and 32 diagnostic right-hand sides), the equilibrated sparse solver reproduces all acceptance sequences, Newton iteration counts, and line-search step sizes of the dense solver. Across 40 parameter directions (the 39 directions in Figure~\ref{fig:msfr_adjoint} plus an all-ones porosity vector), the maximum relative discrepancy between the sparse adjoint and forward tangent-linear sensitivities on the 40-step window is $1.90\times 10^{-10}$ on the archived dense trajectory and $5.74\times 10^{-10}$ on the sparse trajectory.

\subsection{Transient-Adjoint Verification Details}
\label{app:msfr_adjoint}

Table~\ref{tab:msfr_adjoint} summarizes the transient-adjoint verification across the five frozen-schedule trajectory windows plotted in Figure~\ref{fig:msfr_adjoint}. For each parameter direction $\mathbf d$, the independent forward tangent-linear recurrence propagates state sensitivities $\mathbf S_k = d\mathbf y_k/d\epsilon$ via
\begin{equation}
 \mathbf A_k\mathbf S_k=-\left[
 \mathbf R_{k,\boldsymbol\theta}\mathbf d
 +\mathbf R_{k,\mathbf y_{k-1}}\mathbf S_{k-1}
 +\mathbf R_{k,\mathbf y_{k-2}}\mathbf S_{k-2}\right],
 \label{eq:msfr_forward_sensitivity}
\end{equation}
with $\mathbf S_0 = \mathbf 0$.

\begin{table}[htb]
 \centering
 \caption{MSFR transient-adjoint verification across five trajectory windows. ``Max.\ discrepancy'' gives the maximum relative difference between the adjoint and forward tangent-linear recurrence across 39 directions; ``FD counts'' lists resolved-and-agreeing versus unresolved finite-difference directions; and ``Reference change'' gives $e_{\mathrm{stab}}$ (Eq.~\eqref{eq:msfr_gradient_stability}) after applying two additional root-polishing Newton updates per step.}
 \label{tab:msfr_adjoint}
 \small\setlength{\tabcolsep}{4pt}
 \begin{tabular}{lrrr}
 \toprule
 Trajectory window / schedule & Max.\ discrepancy & FD counts & Reference change $e_{\mathrm{stab}}$\\
 \midrule
 1 step, backward Euler & $1.15\times10^{-12}$ & 11/3 & $2.10\times10^{-12}$\\
 3 steps, backward Euler + BDF2 & $6.75\times10^{-12}$ & 11/3 & $5.28\times10^{-12}$\\
 10 steps, with step-9 retry & $7.02\times10^{-12}$ & 11/3 & $1.40\times10^{-11}$\\
 10 steps, baseline schedule & $1.38\times10^{-11}$ & 3/0 & $1.47\times10^{-10}$\\
 40 steps, baseline schedule & $7.37\times10^{-10}$ & 3/0 & $7.93\times10^{-11}$\\
 \bottomrule
 \end{tabular}
\end{table}

To monitor linear-solver accuracy in the presence of an unscaled 1-norm condition number of $\sim 5.5\times 10^{16}$, we track the scaled backward error $\eta_\infty$ and subsequent iterative-refinement step ratio $\delta_{\mathrm{ref}}$:
\begin{equation}
 \eta_\infty=\frac{\|\mathbf M\mathbf x-\mathbf b\|_\infty}
 {\|\mathbf M\|_\infty\|\mathbf x\|_\infty+\|\mathbf b\|_\infty},
 \qquad
 \delta_{\mathrm{ref}}=\frac{\|\Delta\mathbf x\|_\infty}{\|\mathbf x\|_\infty}.
 \label{eq:msfr_linear_diagnostics}
\end{equation}
Across all five windows, the transposed adjoint solves achieve $\eta_\infty \le 1.3\times 10^{-22}$ and $\delta_{\mathrm{ref}} \le 1.5\times 10^{-12}$, while the forward-sensitivity solves achieve $\eta_\infty \le 9.1\times 10^{-17}$ and $\delta_{\mathrm{ref}} \le 5.8\times 10^{-13}$.

Finally, to quantify how finite nonlinear stopping tolerances affect the discrete-adjoint gradient, each operating gradient $\mathbf g^{\mathrm{op}}$ is compared against a root-polished reference gradient $\mathbf g^{\mathrm{ref}}$ computed by forcing two extra Newton updates per time step:
\begin{equation}
 e_{\mathrm{stab}}=\max_j
 \frac{\|\mathbf g_j^{\mathrm{op}}-\mathbf g_j^{\mathrm{ref}}\|_\infty}
 {\|\mathbf g_j^{\mathrm{ref}}\|_\infty}.
 \label{eq:msfr_gradient_stability}
\end{equation}
As shown in the final column of Table~\ref{tab:msfr_adjoint}, $e_{\mathrm{stab}}$ remains between $2.10\times 10^{-12}$ and $1.47\times 10^{-10}$ across all five windows.

\section{Execution Configuration and Compilation Details}
\label{app:execution}

\subsection{GPU Backend Verification and Device Residency}

On the NVIDIA RTX 2000 Ada GPU (PyTorch 2.13.0+cu129, Triton 3.7.1, cuDSS 0.8.0.10, \texttt{float64} precision, fused multiply--add (FMA) contraction disabled), the sparse GPU backend matches the CPU SuperLU implementation on the 40-step and 10-step retry windows within maximum normalized state differences of $1.08\times 10^{-14}$ and $1.26\times 10^{-14}$, and maximum blockwise gradient differences of $1.78\times 10^{-12}$ and $8.91\times 10^{-12}$, respectively.

To reduce kernel launch overhead during element-Jacobian assembly on the GPU, deterministic atomic accumulation is relaxed only on element-assembly passes verified at initialization to have duplicate-free write indices, while global assembly, residual VJPs, and linear solves retain deterministic execution. Combining device convergence checks into a single scalar host read per compiled call reduces host--device scalar reads from 361 to 249 per Newton step and from 128 to 68 per adjoint step, keeping all field tensors, sparse matrix buffers, and trajectory history records resident on the GPU.

\subsection{CPU Repeatability and Compiled-Kernel Agreement}

Under the baseline four-thread CPU configuration (\texttt{MKL\_CBWR=AUTO,STRICT}, fixed 4 Torch/OpenMP/MKL threads, and PyTorch deterministic algorithms enabled), repeated forward and adjoint runs produce bitwise-identical trajectories and gradients across fresh processes and under 24-worker CPU background contention. Table~\ref{tab:msfr_compilation} summarizes the numerical agreement between eager and \texttt{torch.compile}-compiled CPU execution on the dense backend, showing agreement to within $8.35\times 10^{-16}$ on individual kernels and $1.14\times 10^{-13}$ on 40-step adjoint gradients.

\begin{table}[htb]
 \centering
 \caption{Maximum normalized discrepancies between compiled (\texttt{torch.compile}) and eager MSFR execution on CPU (\texttt{float64}, 4 threads).}
 \label{tab:msfr_compilation}
 \small
 \begin{tabular}{lr}
 \toprule
 Evaluated quantity & Maximum normalized discrepancy\\
 \midrule
 Assembled residual vector & $8.35\times10^{-16}$\\
 Residual vector--Jacobian product (VJP) & $4.27\times10^{-16}$\\
 Element-Jacobian blocks & $6.61\times10^{-16}$\\
 Dense LU factors and linear-system solutions & Bitwise identical\\
 Adjoint gradient (40-step window) & $1.14\times10^{-13}$\\
 Adjoint gradient (10-step retry window) & $1.40\times10^{-14}$\\
 \bottomrule
 \end{tabular}
\end{table}

Because the transient uses a backward-Euler startup step followed by variable-step BDF2, \texttt{torch.compile} builds separate graph specializations for the residual, element-Jacobian, and adjoint VJP operations under each time integrator. The first-execution forward costs reported in Figure~\ref{fig:msfr_200_cost}(a) include compilation, setup, and execution of the forward path, rather than compiler time alone. Adjoint VJP specializations are compiled separately when first invoked.

\end{document}